\documentclass[12pt,draftclsnofoot,onecolumn]{IEEEtran}
\usepackage{hyperref}
\usepackage{color}
\usepackage{amsmath}
\usepackage{cite}
\usepackage{bm}
\usepackage{mathrsfs}
\usepackage{multirow}
\usepackage{subfig}
\usepackage{graphicx}
\usepackage{amssymb}
\usepackage{calc}
\usepackage{stfloats}
\usepackage{comment}
\usepackage{algorithm}
\usepackage{algorithmic}
\usepackage{dsfont}
\usepackage[mathlines,switch]{lineno}
\usepackage{multirow}
\usepackage{epstopdf}
\usepackage{textcomp}
\usepackage{lipsum}
\usepackage{float}
\usepackage{threeparttable}
\usepackage{booktabs}
\usepackage{mathtools}
\graphicspath{{images/}}

\usepackage{setspace}%


\DeclareMathAlphabet{\mathpzc}{OT1}{pzc}{m}{it}
\floatname{algorithm}{Module}

\newcommand{\Rev}[1]{{\color{black}{#1}}}

\begin{document}


\title{\Rev{\huge Joint Target Detection and Tracking in Multipath Environment: A Variational Bayesian Approach}}

\author{Hua Lan\thanks{All authors are with the School of Automation, Northwestern Polytechnical University, and the Key Laboratory of Information Fusion Technology, Ministry of Education, Xi'an, Shaanxi Province, 710072, PR China. Hua Lan is also with the School of Electronic Engineering, Xidian University, and the National Laboratory of Radar Signal Processing, Xi'an, Shaanxi Province, 710071, PR China. This work was supported by the National Natural Science Foundation of China (Grant No.~61873211, 61501378, 61503305, 61790552).}, Shuai Sun, Zengfu Wang*\thanks{*~Corresponding author:~Z.~Wang.}, Quan Pan, Zhishan Zhang}

\maketitle

\begin{abstract}
We consider multitarget detection and tracking problem for a class of multipath detection system where one target may generate multiple measurements via multiple propagation paths, and the association relationship among targets, measurements and propagation paths is unknown.
In order to effectively utilize multipath measurements from one target to improve detection and tracking performance, a tracker has to handle high-dimensional estimation of latent variables including target active/dormant meta-state, target kinematic state,
and multipath data association.
Based on variational Bayesian inference, we propose a novel joint detection and tracking algorithm that incorporates multipath data association, target detection and target state estimation in a unified Bayesian framework. The posterior probabilities of these latent variables are derived in a closed-form iterative manner, which is effective for reducing the performance deterioration caused by the coupling between estimation errors and identification errors.
Loopy belief propagation is exploited to approximately calculate the probability of multipath data association, saving the computational cost significantly. Simulation results of over-the-horizon radar multitarget tracking show that the proposed algorithm outperforms multihypothesis multipath track fusion and multi-detection (hypothesis-oriented) multiple hypothesis tracker, especially under low signal-to-noise ratio circumstance.
\end{abstract}


\section{Introduction}\label{sec:introduction}
Multitarget detection and tracking~(MDT)~is essential for many applications in the areas of defense,
medical science, traffic control and navigation \cite{Bar-Shalom2001}.
Most algorithms~\cite{Cox1993,Pulford2005} in the literature addressing MDT are based on the assumption that,
in a single scan, a target generates at most one measurement and one measurement comes from at most one target.
For certain sensing systems, however, one target may result in multiple measurements that are not spatially structured
due to multipath propagation phenomenon \cite{Baum2016}.
Such sensing systems, called multipath detection systems~(MDS) in this paper, include skywave over-the-horizon radar~(OTHR) \cite{Fabrizio2013},
passive coherent location~(PCL) system \cite{Tharmarasa2012}, the sensors work in urban environments \cite{Zhou2016,Li2014}, \Rev{etc}.
In OTHR, high-frequency radiowave propagation through a layered ionosphere often gives rise to multiple propagation paths,
resulting in multiple resolved detections for one target at each radar scan.
Likewise, a PCL with multistatic configurations may also receive multiple detections from various transmitters of opportunity for a single target.
Multiple delayed returns for one target may be obtained when sensors work in urban environments because of dense terrains.

For MDS, if an MDT algorithm extracts target information in multiple measurements properly,
both estimation and detection performance can be improved by the increase of signal-to-noise ratio (SNR) \cite{Xu2015}.
However, as elucidated in the following, unknown measurement-to-path association brings several additional issues,
making MDT for MDS more challenging.
In terms of \textit{target tracking}, (1) measurements need to be associated with tracked targets,
and besides that a target-originated measurement needs to be associated with one of the propagation paths;
(2) a tracker has to handle high-dimensional estimation of \Rev{latent} variables including target active/dormant meta-state, target kinematic state,
and multipath data association; (3) exact computation of multipath data association is intractable due to the ``combinatorial explosion'' in summing out target-to-measurement-to-path association events.
With regard to \textit{target detection}, (1) the number of targets is unknown and time-varying, and apart from that targets may appear and disappear anytime anywhere in the absence of prior information; (2) as many as $N_P$ tracks from $N_P$ propagation paths can be produced for one target.
Additionally, target detection and target tracking are tightly coupled.
Existing work on MDT for MDS can be summarized as the following three categories.

Formulating MDT as two separate and sequential sub-problems, single-path target tracking \cite{Pulford1996} and multipath track fusion \cite{Percival1998}, algorithms in the first category have a simple structure and are computationally efficient.
However, since only measurements from a single path are used in the single-path target tracking procedure,
(multipath)~tracks generated under each path are not accurate or even missed under extreme low SNR circumstance, making fused tracks unreliable.

The second category algorithms address MDT by extending single-path data association to multipath data association,
such as multipath probabilistic data association~(MPDA) \cite{pulford1998multipath, pulford2004othr},
multihypothesis Viterbi data association~(MVDA) \cite{PulfordAES2010},
multi-detection joint probabilistic data association~(MD-JPDA) \cite{Habtemariam2013},
multi-detection multiple hypothesis tracker~(MD-MHT) \cite{sathyan2013multiple},
multi-detection probability hypothesis density~(MD-PHD) \cite{Xu2015},
and multi-detection Bernoulli~(MD-Bernoulli) \cite{Chen2014}.
By performing triple target-to-measurement-to-path association, these algorithms fuse information on one target at the measurement level,
leading to remarkable improvement on tracking performance comparing with the first category algorithms.
One drawback of the second category algorithms is that they are unable to deal with the coupling between data association and target state estimation, which means errors (e.g., a track is by chance associated with clutter or/and a measurement is associated with a wrong path) cannot be corrected once they have been made \cite{Turner2014}.
\Rev{Among the above-mentioned algorithms, MPDA, MVDA and MD-Bernoulli are capable only to track a single target, while MD-JPDA assumes the number of targets is known and fixed. Others including MD-MHT and MD-PHD are able to perform MDT jointly.
MD-MHT approximates the triple target-to-measurement-to-path association by path-dependent two-dimensional assignments and suffers from high computational cost.}
Based on the random finite set theory, MD-PHD is time-consuming as well~\cite{Xu2015}.

The key idea in the third category algorithms is \Rev{to perform} target state estimation and data association
jointly based on expectation-maximization~(EM) framework~\cite{Dempster1977}.
The EM-based algorithms in \cite{Pulford1997EM, Lan2014TSP, Lan2014Fusion} alternate between computing the expected complete log-likelihood according to the posterior probability density function~(PDF) of missing data~(multipath data association) in E-Step and optimizing it with respect to (w.r.t.) the model parameters~(state estimation) in M-Step, which are attractive and desirable to reduce the performance deterioration caused by the coupling between identification errors (from data association) and estimation errors \cite{Lan201652}. However, these work considered a single target tracking.
Extending the work of \cite{Pulford1997EM, Lan2014TSP, Lan2014Fusion} to multitarget tracking with time-varying number of targets would give rise to high-dimensional \Rev{latent} variables, in which case the EM algorithm may suffer from slow convergence \cite{Tzikas2008}.

\Rev{Unlike EM that requires to calculate the conditional expectation of high-dimensional latent variables, variational Bayes~(VB) \cite{Blei2016variational}, as an extension of EM algorithm, casts an inference problem as an optimization problem and provides analytical approximations to the posterior distributions of \Rev{latent} variables, and is superior to EM for complex graphical models~\cite{Bishop2006}.}
VB is gaining popularity within the target tracking community due to its following capabilities and advantages: (1) it provides a unified Bayesian framework for joint detection and tracking; (2) it tackles the coupling issue between estimation of \Rev{latent} variables via an iterative optimization; (3) it avoids the enumeration of association hypotheses by using message passing on graphical models;
(4) it converges under mild conditions \cite{Bishop2006}.
Based on VB, S\"{a}arkk\"{a} and  Nummenmaa \cite{Sarkka2009} proposed an adaptive Kalman filtering method to joint estimation of dynamic state and the time-varying parameters of measurement noise.
Laet \emph{et al.} \cite{Laet2011} proposed an online two-level multitarget tracking and detection algorithm for targets with multiple measurements,
where VB is used in the measurement clustering level.
L\'{a}zaro-Gredilla \emph{et al.} \cite{Lazaro2012} introduced a mixture of Gaussian processes to model the mixed measurements received from multiple targets. The hyperparameters of the mixture Gaussian are learned based on VB.
Orguner \cite{Orguner2012} proposed a VB based method for extended target tracking.
Turner \emph{et al.} \cite{Turner2014} proposed a VB-based tracker integrating track management, data association, and state estimation via an iterative manner, where belief propagation is used to solve data association problem.
Lau \emph{et al.} \cite{Lau2016} proposed a multitarget tracking and detection algorithm based on structured mean-field VB. Williams and Lau~\cite{Willams2016} proposed a multiple scan data association method based on convex fractional free energy. However, to our best knowledge, no one has considered MDT for MDS by using VB.

This paper considers the MDT problem for MDS, and aims to derive the joint posterior distribution over
the high-dimensional \Rev{latent} variables including target kinematic state, target active/dormant meta-state and multipath data association given the received measurements.
By leveraging on \emph{Mean-field} VB, the joint posterior distribution over the high-dimensional \Rev{latent} variables
is approximated by a distribution which is from the family of the product of individual posterior distributions over target kinematic state, target meta-state and multipath data association, and that has the minimal Kullback-Leibler~(KL) divergence to the joint posterior distribution.
The \textit{prior} distribution functions of target kinematic state, target meta-state and multipath data association are constructed from exponential family.
Coordinate ascent together with an iterative mechanism are used to optimize the approximate posterior distributions over target kinematic state, target meta-state and multipath data association.
In each iteration, fixed-interval smoother, forward-backward algorithm are used to estimate target kinematic state and target meta-state, respectively.
Modeled by a factor graph, multipath data association integrates measurements from multiple path at the measurement level and the corresponding association probability is approximated by loopy belief propagation~(LBP).
In summary, the key contributions of this paper are as follows.
\begin{itemize}
\item For the first time, we incorporate multipath data association, target detection and target kinematic state estimation in a unified Bayesian framework for joint MDT of MDS.
\item We propose a computationally efficient joint MDT algorithm for MDS, JDT-VB, which fuses information on one target at the measurement level and is capable of minimizing the performance deterioration caused by the coupling between identification errors and estimation errors, improving target detection and tracking performance for MDS significantly.
\item \Rev{To circumvent the enumeration of all joint multipath data association hypotheses, a factor graph for modeling the multipath data association is presented, and the corresponding marginal association probability is approximated by LBP.}
\end{itemize}
\Rev{Our work in this paper was inspired by the work of \cite{Turner2014} but differs from \cite{Turner2014} in three main aspects.
Firstly, multiple propagation paths are involved in measurement equation, making MDT more challenging.
Secondly, a new technique is proposed for estimating target kinematic state by fusing the path-dependent state estimates.
Thirdly, a new factor graph is used for modeling multipath data association.
}

A preliminary version of the results presented here appeared in two conference papers \cite{lan2016IF,Sun2016IF}.
Here, we present for the first time the entire formulation including implementation details, and also supplement initialization, computational complexity analysis and simulations of the proposed algorithm.

The rest of the paper is organized as follows. The problem formulation of joint MDT for MDS is described in Section~\ref{sec:problemformulation}.
The closed-form analytical solutions of target meta-state estimation, target kinematic state estimation and multipath data association are derived in Section~\ref{sec:solution}.
The simulation analysis and the conclusion are given in Section~\ref{sec:simulation} and Section~\ref{sec:conclusion}, respectively.

\section{Problem Formulation}\label{sec:problemformulation}
Consider the following discrete-time dynamic system
\begin{equation}\label{equ1}
x_{i, k+1} = f_k(x_{i, k}) + v_{i, k + 1}, \quad i = 1, 2,\ldots, N_T,
\end{equation}
where $x_{i, k} \in \mathcal{R}^{n_x}$ is the $i$th target kinematic state to be estimated at time~(scan) $k$ with $n_x$ being the dimension of $x_{i,k}$, $f_k$ is the known kinematic state transition function, and $v_{i, k}$ is a zero-mean Gaussian white noise with known covariance matrix $Q_{i, k}$. Here, $N_T$ is the maximum number of targets in the region of interest, which can be determined by counting all candidate tracks~(targets) in initialization stage. The initial kinematic states of targets $x_{i,0}$, $i=1,2,\ldots, N_T$ are assumed to be Gaussian distributed.

Denote $y_{j, k} \in \mathcal{R}^{n_y}, j = 1, \ldots, N_{k,M}$ as the $j$th measurement at time $k$ with $n_y$ being the dimension of $y_{j,k}$ and $N_{k, M}$ being the known number of measurements. For the MDS, each received measurement $y_{j,k}$ might be originated from an underlying target through a particular propagation path or clutter. The multipath measurement model for the MDS is as follows.
\begin{equation}\label{equ2}
y_{j, k} =
\begin{cases}
h_k^{1}(x_{i, k}) + w_{1, k}    & \text{if $y_{j,k}$ is originated from $i$th target through path $1$}\\
h_k^{2}(x_{i, k}) + w_{2, k}    & \text{if $y_{j,k}$ is originated from $i$th target through path $2$}\\
 \vdots                         & \vdots\\
h_k^{N_P}(x_{i, k}) + w_{N_P, k}& \text{if $y_{j,k}$ is originated from $i$th target through path $N_P$}\\
\text{clutter}                  & \text{otherwise}
\end{cases}
\end{equation}
where $h_k^{\tau}(\cdot)$, $\tau = 1, 2, \ldots, N_P$, is the measurement function of the $\tau$th propagation path with $N_P$ being the known number of paths, and $w_{\tau, k}$ is the corresponding measurement noise, which is assumed to be a zero-mean Gaussian random variable with covariance $R_{\tau, k}$. Here, $v_{i,k}$, $w_{\tau,k}$ and $x_{i,0}$ are assumed to be independent.
As in \cite{Bar-Shalom2001}, the false measurement~(clutter)  is assumed to be uniformly distributed in the region of interest and the number of clutter per scan is assumed to follow a Poisson distribution.
\begin{figure}[!htbp]
	\centering
	\includegraphics[width=0.5\textwidth]{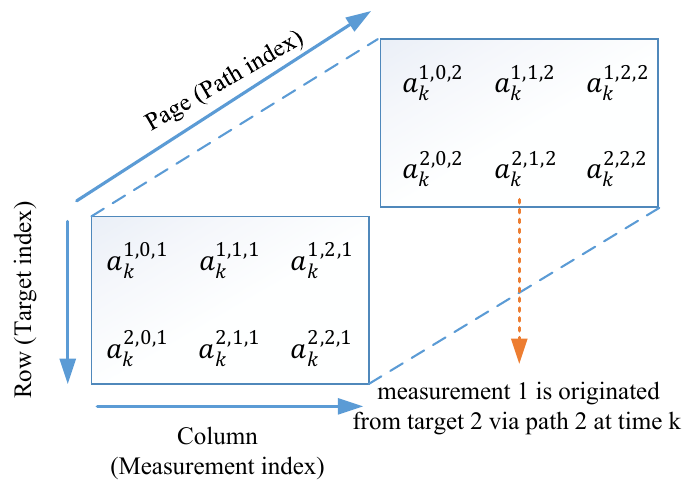}
	\caption{Multipath association events $a_k^{i, j, \tau} (i > 0)$~(two targets, two measurements and two paths).}
	\label{Fig2-a}
\end{figure}

\Rev{
As illustrated by Fig.~\ref{Fig2-a}, the target-to-measurement-to-path association event~(or hypothesis) $a_k^{i,j,\tau}, i > 0, j>0$ represents that the $j$th measurement is originated from the $i$th target via the $\tau$th path.
For $i > 0$, $a_k^{i, 0, \tau}$ means that the $i$th target is not detectable by the $\tau$th path.
We use $a_k^{0, j}, j > 0$ to represent that the $j$th measurement is originated from clutter since a clutter is irrespective of the propagation path.
Let $A_k = \big\{a_k^{i,j,\tau}\big\}_{i=1,j=0,\tau=1}^{N_T, N_{k,M}, N_P} \bigcup \big\{a_k^{0,j}\big\}_{j=1}^{N_{k,M}}$ be a joint event. }
\vspace{-4pt}
By the assumptions that a measurement via any one propagation path has only one source, for each propagation path at most one measurement is received from a target,
and a measurement of a target is received through at most one propagation path \cite{sathyan2013multiple},
the joint event $A_k$ is constrained by the following equation group.
\begin{equation}\label{frame}
\begin{split}
&\sum \limits_{i = 1}^{N_T} a_k^{i, j, \tau} + a_k^{0, j} = 1,   \forall \, j \!=\! 1, \dots, N_{k, M}, \tau \!=\! 1, \dots, N_{P}, \\
&\sum \limits_{j = 0}^{N_{k, M}} a_k^{i, j, \tau} = 1, \quad \forall \, i = 1, \dots, N_{T}, \tau = 1, \dots, N_{P}, \\
&\sum \limits_{\tau = 1}^{N_{P}} a_k^{i, j, \tau} = 1, \quad \forall \, i = 1, \dots, N_{T}, j = 1, \dots, N_{k, M}.
\end{split}
\end{equation}
A joint event $A_k$ is \emph{feasible} if it satisfies equation group~(\ref{frame}).
Note that merged measurements that may be originated from multiple targets via different propagation paths are not considered in this paper.

Define the binary variable $s_{i, k} \in \{0, 1\}$ to represent the active/dormant meta-state of target $i$.
By \emph{active state} of a target we mean that the target is detectable with high probability.
On the contrary, a target is in \emph{dormant state} if it is hardly detectable or it disappears.
Here, $s_{i, k} = 1$ if target $i$ is in active state at time $k$ and $s_{i, k} = 0$ if target $i$ is in dormant state. 
The sets of measurements, target kinematic state, target meta-state are defined by $Y_k = \{y_{j,k}\}_{j=1}^{N_{k, M}}$, $X_k = \{x_{i,k}\}_{i=1}^{N_T}$, $S_k = \{s_{i,k}\}_{i=1}^{N_T}$, respectively.
Note that targets may appear/disappear at different time in the region of interest, i.e., they may have
different lifetime.
For ease of exposition~(and without loss of generality), we hereafter restrict the notation on the lifetime of all targets from $1$ to $K$.
The \Rev{sequences of sets} $Y_{1}^K$, $X_{1}^K$, $S_{1}^K$ and $A_{1}^K$ refer to the collections of measurements, target kinematic state, target meta-state, target-to-measurement-to-path association events up to and including time $K$, respectively.
That is, $Y_{1}^K = \{Y_k\}_{k=1}^K$, $X_{1}^K = \{X_k\}_{k=1}^K$, $S_{1}^K = \{S_k\}_{k=1}^K$, $A_{1}^K = \{A_k\}_{k=1}^K$.
\Rev{In this paper, target state variables~(e.g., $X_{1}^K$, $S_{1}^K$) and data association variables~(e.g., $A_{1}^K$), which are not directly observed but are rather inferred from measurements~(e.g., $Y_{1}^K$), are called \emph{latent variables}.}

Based on the models of target kinematic state, \Rev{target meta-state}, multipath measurement and multipath data association, our problem of joint MDT for MDS is to concurrently estimate target kinematic state $X_{1}^K$ (tracking) and target meta-state $S_{1}^K$ (detection),
given measurements $Y_{1}^K$ in the presence of unknown multipath data association $A_{1}^K$, i.e.,
\begin{equation}\label{Joint}
\begin{split}
p(X_1^K|Y_{1}^K) &= \sum \nolimits_{A_1^K} p(X_1^K|Y_1^K, A_1^K) p(A_1^K), \\
p(S_{1}^K|Y_1^K) &= \sum \nolimits_{A_1^K} p(S_1^K|Y_1^K, A_1^K) p(A_1^K).
\end{split}
\end{equation}
%

A common way to solve the above-stated joint MDT problem~(\Rev{Eq.~(\ref{Joint})}) is to calculate joint posterior PDF $\mathcal{L}_1^K := p(X_{1}^K, S_{1}^K, A_{1}^K|Y_{1}^K)$ first, and then marginalize the joint posterior PDF $\mathcal{L}_1^K$ to obtain the posterior PDF of target kinematic state $X_{1}^K$ and posterior probability mass function~(PMF)~of target meta-state $S_{1}^K$.
Since the number of multipath data association events increases exponentially with the number of targets, the number of (validated) measurements and the number of paths, exact calculation of $\mathcal{L}_1^K$ is computationally expensive and approximation solutions are often sought.
The sampling-based stochastic approximation methods~(e.g., MCMC) are computationally intensive~\cite{Blei2016variational}.
VB, as an analytical-based deterministic approximation method, is more computationally effective and suited to MDT problem of MDS.

\section{VB for Joint Detection and Tracking}\label{sec:solution}
VB provides a local-optimal, exact analytical solution to the approximation of an inference problem, making it
more competitive in the case of high-dimensional \Rev{latent} variables.
In this paper, we employ VB to solve the joint MDT of MDS.

The first step in applying VB is to pick up a simpler family of probability densities over the \Rev{latent} variables $Z_{1}^K = \{X_{1}^K, S_{1}^K, A_{1}^K\}$.
We will specify the form of the family we use later.
Then, we seek the distribution $q(Z_{1}^K)$ closest to the posterior PDF $p(Z_{1}^K|Y_{1}^K)$ in this family and use it to
approximate $p(Z_{1}^K|Y_{1}^K)$, i.e., $p(Z_{1}^K|Y_{1}^K) \approx q(Z_{1}^K)$.
We choose the commonly used Kullack-Leibler~(KL)~divergence as the dissimilarity function of two distributions.
The KL-divergence from the density $q(Z_{1}^K)$ to the density $p(Z_{1}^K|Y_{1}^K)$ is defined as
\begin{equation}\label{eq:KL}
\begin{split}
\text{KL}(q||p) & = \mathbb{E}_{q(Z_{1}^K)} \log q(Z_{1}^K) - \mathbb{E}_{q(Z_{1}^K)} \log p(Z_{1}^K | Y_{1}^K) \\  & = \log p(Y_{1}^K) - \mathcal{B}(Z_{1}^K),
\end{split}
\end{equation}
where
\begin{equation}\label{eq:BK}
\mathcal{B}(Z_{1}^K) = \mathbb{E}_{q(Z_{1}^K)} \log p(Z_{1}^K, Y_{1}^K) - \mathbb{E}_{q(Z_{1}^K)} \log q(Z_{1}^K)
\end{equation}
is known as the variational free energy, which is also the lower bound for the evidence $\log p(Y_{1}^K)$ since the $\text{KL}(q||p)$ is nonnegative.
Evidently, minimizing $\text{KL}(q||p)$ is equivalent to maximizing $\mathcal{B}(Z_{1}^K)$.

We use \emph{mean-field variational family} \cite{Blei2016variational} where $Z_{1}^K$ is partitioned into three disjoint groups
$X_{1}^K$, $S_{1}^K$, $A_{1}^K$ and each of the factors $q(X_{1}^K)$, $q(S_{1}^K)$, $q(A_{1}^K)$ is a probability distribution with a free functional form.
The joint distribution over high-dimensional \Rev{latent} variables are represented by the product of individual PDFs~(or PMF) of the low-dimensional \Rev{latent} variable. That is,
\begin{equation}\label{eq:QK}
q(Z_{1}^K) = q(X_{1}^K) q(S_{1}^K) q(A_{1}^K).
\end{equation}

In general, $\mathcal{B}(Z_{1}^K)$ is non-convex and no explicit solution for $q(Z_1^K)$ can be found.
Coordinate ascent, a computationally effective technique with competitive performance for large-scale non-convex machine learning problems,
is exploited to \Rev{maximize} Eq.~(\ref{eq:BK}).
Next we show how to apply coordinate ascent to solve Eq.~(\ref{eq:BK}).

Substitute Eq.~(\ref{eq:QK}) into Eq.~(\ref{eq:BK}), and rewrite Eq.~(\ref{eq:BK}) as a function of the factor $q(X_{1}^K)$ for example,
one obtains,
\begin{equation}\label{eq:bz1k}
\begin{split}
\mathcal{B}(Z_{1}^K) = \mathbb{E}_{q(X_1^K)} \mathbb{E}_{q(S_1^K), q(A_1^K)} \log p(Z_1^K, Y_1^K)
- \mathbb{E}_{q(X_1^K)} \log q(X_1^K) + c,
\end{split}
\end{equation}
where $c = -\mathbb{E}_{q(S_1^K)} \log q(S_1^K) - \mathbb{E}_{q(A_1^K)} \log q(A_1^K)$, a constant w.r.t. $X_{1}^{K}$.
Let
\begin{equation}\label{eq:QX}
\begin{split}
q^*(X_{1}^K) & = \alpha \exp \left(\mathbb{E}_{q(S_{1}^K), q(A_{1}^K)} \log p(Z_{1}^K, Y_{1}^K) \right) \\
& \propto \exp \left(\mathbb{E}_{q(S_{1}^K), q(A_{1}^K)} \log p(Z_{1}^K, Y_{1}^K) \right),
\end{split}
\end{equation}
where $\alpha$ is the normalization constant.

Substituting Eq.~(\ref{eq:QX}) into Eq.~(\ref{eq:bz1k}) and by the definition of KL divergence, we have
\begin{equation}
\mathcal{B}(Z_{1}^K) = - \text{KL}\left( q(X_{1}^K) || q^*(X_{1}^K) \right ) + c - \log \alpha.
\end{equation}

Clearly, $\mathcal{B}(Z_{1}^K)$ takes the maximum value when $q(X_{1}^K) = q^*(X_{1}^K)$.
Similarly,
\begin{equation}\label{eq:QSA}
\begin{split}
q^*(S_{1}^K) &\propto \exp \left(\mathbb{E}_{q(X_{1}^K), q(A_{1}^K)} \log p(Z_{1}^K, Y_{1}^K) \right),\\
q^*(A_{1}^K) &\propto \exp \left(\mathbb{E}_{q(X_{1}^K), q(S_{1}^K)} \log p(Z_{1}^K, Y_{1}^K) \right).
\end{split}
\end{equation}
Note that $q^*(X_{1}^K)$, $q^*(S_{1}^K)$, $q^*(A_{1}^K)$ are not explicit solutions since
calculation of any $q^*(\cdot)$ depends on expectations computed w.r.t. the other two factors.
Therefore, the iterative mechanism, which starts from appropriately initialized $q^*(\cdot)$ and then cycles through the factors and update each in turn by recalculating it using the current values for the other two factors, is resorted to estimating $q(\cdot)$.
Since $\mathcal{B}(Z_{1}^K)$ is convex w.r.t. each of the factors, the eventually derived estimation of the factors are local optimal \cite{Bishop2006}.

Based on the principled application of VB described above, we next propose a joint MDT algorithm named JDT-VB for MDS.
In the following, we first present the framework of the proposed JDT-VB algorithm,
and then provide all the implementation details.

\subsection{Framework of JDT-VB Algorithm}
\begin{figure}[!htbp]
	\vspace{-15pt}
	\centering
	\includegraphics[scale = 0.85]{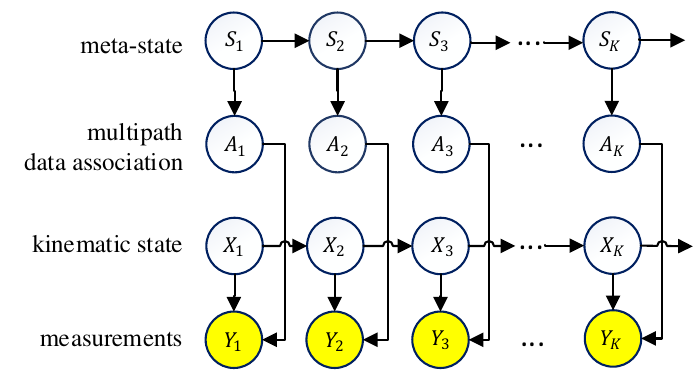}
	\caption{Graphical model of joint detection and tracking.}
	\label{Fig3}
\end{figure}

In the same vein of \cite{Turner2014}, the interdependence among the variables in the joint MDT problem can be described by the graphical model shown in Fig.~\ref{Fig3}.
Target kinematic state $X_k$ and target meta-state $S_k$ evolve with first-order Markov process.
Multipath data association $A_k$ is assumed to be independent over time.
At each time, observable variable $Y_k$ is generated from $X_k$ via a particular path or clutter, and the relationship among target-to-measurement-to-path association is represented by $A_k$.
Additionally, $A_k$ is related to target meta-state $S_k$.
\Rev{Given multipath data association $A_1^K$, the measurements $Y_1^K$ are independent from target meta-state $S_1^K$. The joint PDFs can be factorized as
\begin{equation}\label{13}
p(Y_1^K, X_1^K, A_1^K, S_1^K) = p(Y_1^K | X_1^K, A_1^K) p(A_1^K|S_1^K) p(S_1^K) p(X_1^K).
\end{equation}}

\begin{figure}[!htbp]
	\centering
	\includegraphics[width=0.65\textwidth]{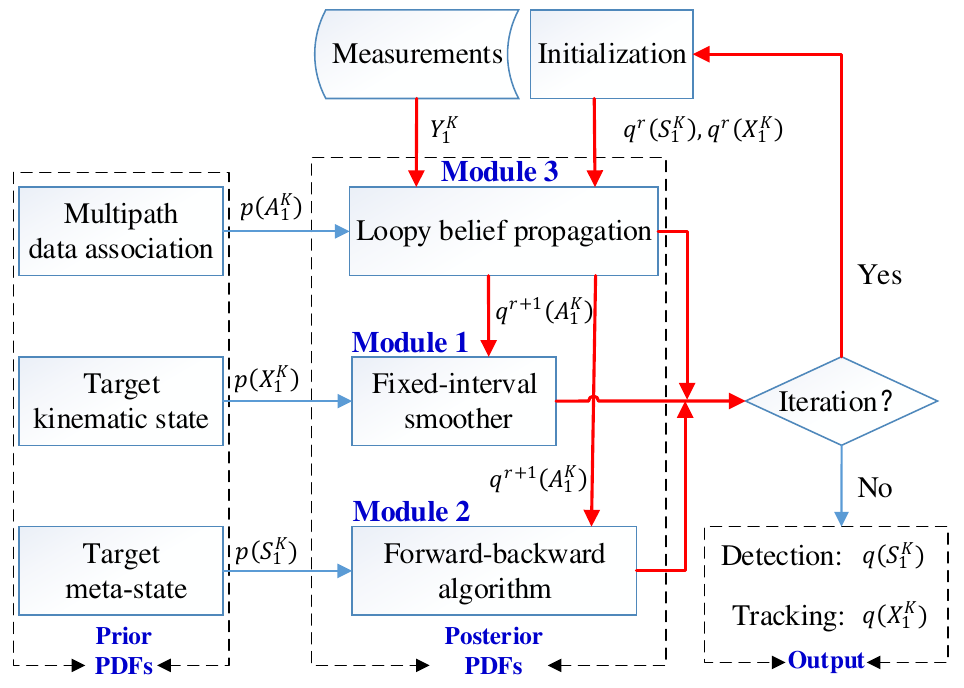}
	\caption{Framework of JDT-VB algorithm.}
	\label{Fig4}
\end{figure}

Based on the graphical model in Fig.~\ref{Fig3},
the framework of the proposed JDT-VB algorithm is depicted in Fig.~\ref{Fig4}.
After constructing the prior probabilities of \Rev{latent} variables,
the posterior PDFs or PMF of target kinematic state, target meta-state, and multipath data association are updated in an iterative loop.
The details are further explained below.
\begin{itemize}
\item  \textit{Prior PDFs modeling}: one convenient and favorable class of priors is conjugate priors in the exponential family. In this paper, the prior PDFs~(or PMFs) of target kinematic state, target meta-state and multipath data association are assumed to be a Gaussian distribution, a Bernoulli distribution, and a distribution in exponential family composed of a Poisson distribution and a Bernoulli distribution, respectively.
\item  \textit{Posterior multipath data association update (Module \ref{alg3})}: given the $r$th iterative (path-unconditional) PDF of target kinematic state $q^r(X_1^K)$ and PMF of target meta-state $q^r(S_1^K)$, the PDF of multipath data association $q^r(A_1^K)$ belongs to exponential family. The distribution of each multipath data association event is approximately calculated by the LBP method.
\item  \textit{Posterior target kinematic state update (Module \ref{alg1})}: given measurements $Y_1^K$ and the $r$th iterative PDF of multipath data association $q^r(A_1^K)$, the PDF of target kinematic state updated by measurements from a single path, called \emph{path-conditional} PDF, is a Gaussian distribution, of which the mean and covariance are obtained via a fixed-interval smoother. The PDF of target kinematic state updated by measurement from all paths, called \emph{path-unconditional} PDF and denoted by $q^r(X_1^K)$, is a Gaussian distribution as well, and is obtained by fusing the path-conditional PDFs of target kinematic state.
\item  \textit{Posterior target meta-state update (Module \ref{alg2})}: given the $r$th iterative PDF of multipath data association $q^r(A_1^K)$, the PMF of target meta-state $q^r(S_1^K)$ is updated via forward-backward algorithm, which is able to integrate statistical information~(e.g., detection probability, false alarm rate) relevant to the specific application.
\item \textit{Iterative loop}: Module 1- Module 3 above are repeated until $\mathcal{B}(Z_1^K)$ at two consecutive iterations are close enough or the maximum number of iterations is reached.
\end{itemize}

\subsection{The Conjugate Prior PDFs and Observation Modeling}
\subsubsection{Target Kinematic State Modeling}
As seen from Fig~\ref{Fig3}, each target is assumed to independently follow a first-order Markov process, that is,
\begin{equation}
p(X_1^K) = \prod_{i = 1}^{N_T} p(x_{i, 1}) \prod_{k=2}^{K} p(x_{i, k}|x_{i, k-1}),
\end{equation}
where $p(x_{i, k}|x_{i, k-1}) = \mathcal{N}(x_{i, k}|f_k(x_{i, k-1}), Q_{i, k})$.
\subsubsection{Target meta-state Modeling}
The meta-state of each target is assumed to independently follow a two-state Markov model as shown in Fig~\ref{Fig3},
\begin{equation} \label{8-a}
p(S_1^K) = \prod_{i = 1}^{N_T} p(s_{i, 1}) \prod_{k=2}^{K} p(s_{i, k}|s_{i, k-1}),  s_{i, k} \in \{0, 1\},
\end{equation}
The initial probability $p(s_{i, 1})$ and transition probabilities $T_s(e,c) = p(s_{i, k} = e | s_{i, k-1} = c), \forall e, c \in \{0, 1\}$ are assumed to be known.
The detection decision on target $i$ is made based on $p(s_{i,1:K})$.
%
\subsubsection{Multipath Data Association Modeling}
The PDF of multipath data association $A_k$ consisting of a Poisson distribution to model clutter
and a Bernoulli distribution to model target detection, is given by
\begin{equation} \label{9-E}
\begin{split}
p(A_k| N_{k,M}) = \dfrac{(\lambda V_k)^{N_{C}(A_k)}}{N_{k,M}!}\exp(-\lambda V_{k})
 \prod_{i=1}^{N_T}\prod_{\tau = 1}^{N_P}(p_d^{\tau}(i))^{d_k^{i, \tau}(A_k)} (1 - p_d^{\tau}(i))^{1-d_k^{i, \tau}(A_k)},
\end{split}
\end{equation}
where
\begin{equation}\label{Nc}
N_{C}(A_k) = N_{k,M} - \sum_{i = 1}^{N_T}\sum_{\tau = 1}^{N_P} d_k^{i, \tau}(A_k)
\end{equation}
is the number of measurements that are not associated with any target in the joint event $A_k$,
\Rev{$d_k^{i, \tau}(A_k)$} is the path-dependent detection indicator of target $i$ that indicates whether a measurement
is associated with target $i$ via path $\tau$ in $A_k$, $\lambda$ represents the clutter density, $p_d^{\tau}(i)$ is the detection probability of target $i$  via path $\tau$, and $V_{k}$ represents the volume of the region of interest,
which is the volume of the validation region when gating technique is used~\cite{Bar-Shalom2001}.
Note that the detection probability of target $i$ via path $\tau$, $p_d^{\tau}(i)$, is a function of target active/dormant meta-state $s_{i, k}$. That is, $p_d^{\tau}(i)$ takes a relatively large value if target $i$ is in the active state~($s_{i,k} = 1$), and vice versa.
See \textbf{Appendix A} for the detailed derivation of Eq.~(\ref{9-E}).

\subsubsection{Observation Modeling}
\Rev{
Based on the assumption that measurements are independently distributed conditioned on target kinematic state and multipath data association events~\Rev{(see the graphical model in Fig.~\ref{Fig3})}, the likelihood function $p(Y_1^K|X_1^K, A_1^K) $ can be factorized as
\begin{equation}\label{10}
\begin{split}
p(Y_1^K |X_1^K, A_1^K)  = \prod \limits_{k=1}^{K} p \left( y_{1,k}, \ldots, y_{N_{k,M},k}| x_{1,k}, \ldots, x_{N_T,k}, A_k \right ).
\end{split}
\end{equation}
Recall that $A_k$ is a joint feasible data association event. 
If $a_k^{0,j} = 1$, $j \in \{1, \ldots, N_{k,M} \}$, that is, no target exists, then measurement $j$ is from clutter and 
\begin{equation}\label{notarget}
p(y_{j,k} | a_k^{0,j} = 1) = V_{k}^{-1}
\end{equation}
by the assumption that the clutter is uniformly distributed in the region of interest~\cite{Bar-Shalom2001}.
If $a_k^{i,0,\tau} = 1$, $i \in \{1, \ldots, N_T \}, \tau \in \{ 1, \ldots, N_P \}$, that is, target $i$ is not detected via path $\tau$,
then
\begin{equation}\label{nomeasurment}
p(\emptyset | x_{i,k}, a_k^{i,0,\tau} = 1) = 1.
\end{equation}
Otherwise, if $a_k^{i,j,\tau} = 1$, $i \in \{1, \ldots, N_T \}, j \in \{1, \ldots, N_{k,M} \}$, $\tau \in \{ 1, \ldots, N_P \}$, that is, measurement $j$ is originated from target $i$ through propagation path $\tau$, we have
\begin{equation}\label{normal}
p(y_{j, k}|x_{i, k}, a_k^{i,j,\tau} = 1) = \mathcal{N}(y_{j,k}| h_k^{\tau}(x_{i, k}), R_{\tau, k}).
\end{equation}
By Eqs.~(\ref{notarget}), (\ref{nomeasurment}), (\ref{normal}) and expanding the right side of Eq.~(\ref{10}), likelihood function (\ref{10}) is rewritten as
\begin{equation}\label{113-a}
\begin{split}
 p(Y_1^K|X_1^K, A_1^K) = \prod \limits_{k=1}^{K} \prod \limits_{j=1}^{N_{k, M}} V_k^{-a_k^{0, j}}\!\prod \limits_{i=1}^{N_T}\prod_{\tau =1}^{N_P} \mathcal{N}(y_{j,k}| h_k^{\tau}(x_{i, k}), R_{\tau, k})^{a_k^{i,j,\tau}}.
\end{split}
\end{equation}
Note that observation likelihood $p(Y_k|X_k, A_k)$ represents the likelihood of the measurements $Y_k$ for the given target states $X_k$ and a joint feasible data association event $A_k$ that is an assignment among measurements, targets and propagation paths.
The clutter density $\lambda$ is not involved in $p(Y_k|X_k, A_k)$.
}

\subsection{Posterior Probability Update of JDT-VB}
Based on the conditional independence of approximate distributions imposed by mean-field method,
the approximate posteriors on target kinematic state and target meta-state are factorized over tracks,
and approximate posteriors on multipath data association is factorized over time, that is,
\begin{eqnarray}\label{18}
&q(X_1^K) = \prod \limits_{i = 1}^{N_T} q(x_{i,1:K}),\;\; q(S_1^K) = \prod \limits_{i = 1}^{N_T} q(s_{i,1:K}),\\
\label{18-a}&q(A_1^K) = \prod \limits_{k = 1}^{K} q(A_k).
\end{eqnarray}

\subsubsection{Calculation of $q(X_1^K)$}\label{sec:calqx}
\Rev{
According to the definition of Eq.~(\ref{eq:QX}) and the factorization in Eq.~(\ref{18}),
the update of posterior PDF of each track can be derived separately as follows.
\begin{equation}\label{eq:qx}
\begin{split}
\log q(x_{i,1:K}) \propto \mathbb{E}_{q(s_{i, 1:K}), q(A_{1}^K)} \log p(x_{i,1:K}, A_1^K, s_{i, 1:K}, Y_{1}^K).
\end{split}
\end{equation}
In the vein of Eq.~(\ref{13}), the PDF $p(x_{i,1:K}, A_1^K, s_{i, 1:K}, Y_{1}^K)$ is decomposed as
\begin{equation}\label{eq:pxjoint}
\begin{split}
p(x_{i,1:K}, A_1^K, s_{i, 1:K}, Y_{1}^K) = p(Y_1^K | x_{i,1:K}, A_1^K)
p(A_1^K|s_{i, 1:K}) p(s_{i, 1:K}) p(x_{i,1:K}).
\end{split}
\end{equation}
Substituting Eq.~(\ref{eq:pxjoint}) into Eq.~(\ref{eq:qx}), and by
\begin{equation}
\begin{split}
 \mathbb{E}_{q(s_{i, 1:K}), q(A_{1}^K)} \log p(x_{i,1:K})  &= \log p(x_{i,1:K}), \\ \nonumber
\mathbb{E}_{q(s_{i, 1:K}), q(A_{1}^K)}  \log p(Y_1^K | x_{i,1:K}, A_1^K)
 &= \mathbb{E}_{q(A_1^K)} \log p(Y_1^K|x_{i, 1:K}, A_1^K),
\end{split}
\end{equation}
yield
\begin{equation}\label{eq:qx2}
\begin{split}
\log q(x_{i,1:K}) =  \log  p(x_{i, 1:K}) + \mathbb{E}_{q(A_1^K)} \log p(Y_1^K|x_{i, 1:K}, A_1^K) + c_x^1,
\end{split}
\end{equation}
where $c_x^1 = \mathbb{E}_{q(s_{i, 1:K}), q(A_{1}^K)}\log p(A_{1}^K|s_{i, 1 : K})p(s_{i, 1 : K})$ is independent with $x_{i,1:K}$.
By a similar derivation of Eq.~(\ref{113-a}), the expectation $\mathbb{E}_{q(A_1^K)} \log p(Y_1^K|x_{i, 1:K}, A_1^K)$ is expanded as
\begin{equation}\label{eq:qa1k}
\begin{split}
& \mathbb{E}_{q(A_1^K)} \log p(Y_1^K|x_{i, 1:K}, A_1^K) = \sum \limits_{k=1}^{K} \sum \limits_{j=1}^{N_{k, M}} \sum_{\tau =1}^{N_P}\mathbb{E}\left[ a_k^{i,j,\tau} \right] \log \mathcal{N}(y_{j,k}| h_k^{\tau}(x_{i, k}), R_{\tau, k}) +  c_x^2,
\end{split}
\end{equation}
where $c_x^2 = -\sum_{k=1}^{K} \sum_{j=1}^{N_{k, M}} \mathbb{E}[a_k^{0, j}] \log(V_k)$ is independent with $x_{i, 1:K}$.}

\Rev{
Substituting Eq.~(\ref{eq:qa1k}) into Eq.~(\ref{eq:qx2}), taking the exponential of both sides of Eq.~(\ref{eq:qx2}), and throwing away the terms ($c_x^1$ and $c_x^2$) that do not depend on $x_{i, 1:K}$, yield
\begin{equation}\label{02}
\begin{split}
q(x_{i, 1:K}) \propto  p(x_{i, 1:K})
\prod \limits_{k=1}^K \prod \limits_{j = 1}^{N_{k, M}} \prod \limits_{\tau = 1}^{N_P}  \mathcal{N}(y_{j, k}|h_k^{\tau}(x_{i,k}), R_{\tau,k})^{\mathbb{E}\left[a_k^{i, j, \tau}\right]}.
\end{split}
\end{equation}
By applying the product of Gaussian PDFs~(see details in \textbf{Appendix B}), we obtain
\begin{equation} \label{B1002}
\begin{split}
\prod \limits_{j = 1}^{N_{k, M}} \mathcal{N}(y_{j, k}|h_k^{\tau}(x_{i,k}), R_{\tau, k})^{\mathbb{E}[a_k^{i, j, \tau}]}  = \mathcal{N}(\bar y_{i,\tau, k}|h_k^{\tau}(x_{i,k}), \bar R_{i, \tau, k}),
\end{split}
\end{equation}
where
\begin{equation}\label{eq:synold}
\bar y_{i,\tau,k} = \dfrac{\sum_{j=1}^{N_{k,M}} \mathbb{E}[a_k^{i,j,\tau}] y_{j, k}}{\sum_{j=1}^{N_{k,M}} \mathbb{E}[a_k^{i,j,\tau}]}, \quad
\bar R_{i, \tau, k} = \dfrac{R_{\tau, k}}{\sum_{j=1}^{N_{k,M}} \mathbb{E}[a_k^{i,j,\tau}]}
\end{equation}
are synthetic measurement and its corresponding measurement covariance.
Substituting Eq.~(\ref{B1002}) into Eq.~(\ref{02}), yields,
\begin{equation}\label{B1003}
q(x_{i, 1:K}) \!\propto p(x_{i, 1:K})  \prod \limits_{k=1}^K \prod \limits_{\tau = 1}^{N_P} \mathcal{N}(\bar y_{i, \tau, k}|h_k^{\tau}(x_{i,k}), \bar R_{i, \tau, k}).
\end{equation}}
\Rev{
From Eq.~(\ref{B1003}), it is seen that $q(x_{i, 1:K})$ is equivalent to a dynamical system with multiple (independent) synthetic measurements $\bar y_{i, \tau, k}, \tau = 1, \ldots, N_P$ and corresponding measurement covariance $\bar R_{i, \tau, k}, \tau = 1, \ldots, N_P$.
Note that since for different propagation path $\tau$, the measurement functions $h^{\tau}(\cdot), \tau = 1, \ldots, N_P$ are different,
the synthetic measurements $\bar y_{i, \tau, k}, \tau = 1, \ldots, N_P$ cannot be synthesized further over all paths.
Here, we approximate Eq.~(\ref{B1003}) by fusing path-dependent kinematic state PDFs which are obtained by Kalman smoother for linear system and URTS for nonlinear system~\cite{Simo2013} based on the synthetic measurements corresponding to the path, that is,
\begin{equation}
\label{22-c}
\hat x_{i, k} = P_{i, k} \sum \limits_{\tau = 1}^{N_P} P_{i, \tau, k}^{-1} \hat x_{i, \tau, k}, \,\,
P_{i, k}^{-1} =  \sum \limits_{\tau = 1}^{N_P} P_{i, \tau, k}^{-1},
\end{equation}
where $\hat x_{i, \tau, k} = \mathbb{E}[x_{i, \tau, k}|\bar y_{i, \tau, 1:K}]$, $P_{i, \tau, k} = \mathrm{cov}[\hat x_{i, \tau, k}| \bar y_{i, \tau, 1:K}]$
are the mean and covariance of path-dependent kinematic state PDFs.
In Eq.~(\ref{22-c}), the interdependency of the path-dependent kinematic state PDFs are neglected.}

In order to reduce computational cost, gating technique~\cite{Bar-Shalom2001} can be applied to eliminate very unlikely association between targets and measurements.
Accordingly, the candidate measurements for posterior PDF update of target $i$ under path $\tau$ should fall into the relevant validation region, which is defined as follows \cite{Bar-Shalom2001}:
\begin{equation}\label{21-a}
\Omega_{i, \tau, k}(\gamma) :=  \{y_{j,k} \in Y_k: \mathcal{D}(y_{j,k} - y_{i, \tau, k}^{-}, S_{i, \tau, k}) \leq \gamma \}.
\end{equation}
In Eq.~(\ref{21-a}), for current iteration $r$, measurement prediction $y_{i, \tau, k}^{-} = h_k^{\tau}(\hat x_{i, k}^{r-1})$ and the associated innovation covariance $S_{i, \tau, k} = R_{\tau,k} + H_k^{\tau}P_{i, k}^{r-1}(H_k^{\tau})^T$,  where $\hat x_{i, k}^{r-1}$, $P_{i, k}^{r-1}$ are the estimated target kinematic state and the
corresponding covariance in the last iteration $r-1$, respectively. $H_k^{\tau}$ is the Jacobian matrix of $h_k^{\tau}$ w.r.t. $\hat x_{i, k}^{r-1}$.
$\mathcal{D}$ represents Mahalanobis distance.
The constant $\gamma$ is chosen to make gate probability equal to $p_g$. The volume of the validation region is \cite{Bar-Shalom2001}
\begin{equation}
V_{k}^{i, \tau} = c_{n_y}\gamma^{n_y/2}|S_{i, \tau, k}|^{1/2},
\end{equation}
where the coefficient $c_{n_y}$ depends on the measurement dimension $n_y$. For $n_y = 3$, $c_{n_y} = 4\pi/3$.

The validation region for multipath measurements is then defined to be the union of the multipath validation regions for all targets.
Calculating the volume $V_{k}$ of the validation region for multipath measurements, required by the calculation of data association probability, is complicated since the multipath validation region for each target may overlap~\cite{pulford1998multipath}.
As in \cite{pulford1998multipath}, here we approximate the volume $V_{k}$ by
\begin{equation}\label{V-all}
V_{k} \approx \sum \limits_{i = 1}^{N_T} \max \left\{V_{k}^{i, 1}, \ldots, V_{k}^{i, N_P}\right\}.
\end{equation}
\Rev{
Consequently, the synthetic measurement $\bar y_{i, \tau, k}$ and its corresponding measurement covariance $\bar R_{i, \tau, k}$ in Eq.~(\ref{eq:synold})
can be rewritten as
\begin{eqnarray}
\label{21}\bar y_{i,\tau,k} = \dfrac{\sum_{j: y_{j,k} \in \Omega_{i, \tau, k}(\gamma)} \mathbb{E}[a_k^{i,j,\tau}] y_{j, k}}{\sum_{j: y_{j,k} \in \Omega_{i, \tau, k}(\gamma)} \mathbb{E}[a_k^{i,j,\tau}]},  \quad
\bar R_{i, \tau, k} = \dfrac{R_{\tau, k}}{\sum_{j: y_{j,k} \in \Omega_{i, \tau, k}(\gamma)} \mathbb{E}[a_k^{i,j,\tau}]}.
\end{eqnarray}}

The calculation of $q(X_1^K)$ for the case of linear dynamic system is summarized in \textbf{Module \ref{alg1}}.
\begin{algorithm}
	\caption{\textit{Tracking}: calculation of $q(X_1^K)$ (for linear system)}
	\label{alg1}
	\begin{algorithmic}[1]
		\REQUIRE measurements $Y_1^K$; $\mathbb{E}[A_1^K]$ from \textbf{Module \ref{alg3}};
		\ENSURE $\big\{\hat{X}_1^K, P_1^K\big\}$: target kinematic state;
		\FOR{each target $i = 1 : N_T$}
		\FOR{each path $\tau = 1 : N_P$}
		\FOR{each time $k = 1 : K$}
		\STATE  Select measurements subset $\Omega_{i, \tau, k}$ via Eq.~(\ref{21-a});
		\STATE Calculate $\bar y_{i,\tau,k}$ and $\bar R_{i, \tau, k}$ via Eq.~(\ref{21});
		\ENDFOR
		\STATE Calculate $\{\hat x_{i, \tau, 1:K}, P_{i, \tau, 1:K}\}$ via KS as follows.
		\FOR{each time $k = 1 : K$}
        \STATE $x_{i, \tau, k}^{-} = F_{k-1} \hat x_{i, \tau, k-1}$, \\
	   $P_{i, \tau, k}^{-} = F_{k-1} P_{i, \tau, k-1} F_{k-1}^T + Q_{i,k}$, \\
	   $S_{i, \tau, k} = H^{\tau}_{k}P_{i, \tau, k}^{-} (H_{k}^{\tau})^T + \bar R_{i, \tau, k}$, \\
	   $\mathcal{K}_{i, \tau, k} = P_{i, \tau, k}^{-}(H_{k}^{\tau})^T S_{i, \tau, k}^{-1}$,  \\
	   $\hat x_{i, \tau, k} = x_{i, \tau, k}^{-} + \mathcal{K}_{i, \tau, k}(\bar y_{i,\tau,k} - H_{k}^{\tau} x_{i, \tau, k}^{-})$, \\
	   $P_{i, \tau, k} = P_{i, \tau, k}^{-} - \mathcal{K}_{i, \tau, k}S_{i,\tau, k}\mathcal{K}_{i, \tau, k}^T$.
		\ENDFOR
	\STATE Let $\hat x_{i, \tau, K|K} = \hat x_{i, \tau, K}$, and $P_{i, \tau, K|K} = P_{i, \tau, K}$.
		\FOR{each time $k = K-1 : 1$}
		\STATE $\mathcal{G} = P_{i, \tau, k} F_k^T(P_{i, \tau, k+1}^{-})^{-1}$,  \\
			$\hat x_{i, \tau, k|K} = \hat x_{i, \tau, k} + \mathcal{G}(\hat x_{i, \tau, k+1|K} - \hat x_{i, \tau, k+1})$, \\
			$P_{i, \tau, k|K} = P_{i, \tau, k} + \mathcal{G}(P_{i, \tau, k+1|K} - P_{i, \tau, k+1})\mathcal{G}^T$.
		\ENDFOR
		\ENDFOR
		\FOR{each time $k = 1 : K$}
		\STATE Calculate $\{\hat x_{i, k}, P_{i, k}\}$ via Eq.~(\ref{22-c}).
		\ENDFOR
		\ENDFOR
	\end{algorithmic}
\end{algorithm}

\subsubsection{Calculation of $q(S_1^K)$}
By Eq.~(\ref{eq:QSA}), Eq.~(\ref{18}) and the dependence of multipath data association on target meta-state shown in Fig.~\ref{Fig3}, for each target $i$, we have\Rev{
\begin{equation}\label{24}
\begin{split}
\log q(s_{i,1:K})  = \log p(s_{i,1:K}) +  \mathbb{E}_{q(A_1^K)}\log p(A_1^K|s_{i, 1:K})  + \underbrace{\mathbb{E}_{q(X_1^K), q(A_1^K)} \log p(Y_1^K|X_1^K, A_1^K)}_{\text{term that is independent with $s_{i, 1:K}$}}.
\end{split}
\end{equation}}
Substituting Eq.~(\ref{9-E}) into Eq.~(\ref{24}), yields,
\begin{equation}\label{26-a}
\begin{split}
q(s_{i, 1:K}) \propto p(s_{i, 1})  \exp(\xi_{i, 1}(s_{i, 1})) \prod \limits_{k = 2}^K p(s_{i, k}|s_{i, k-1}) \exp(\xi_{i, k}(s_{i, k}))
\end{split}
\end{equation}
with
\begin{equation}\label{26-b}
\begin{split}
\xi_{i,k}(s_{i,k})  = \sum_{\tau = 1}^{N_P} \mathbb{E}[d_k^{i, \tau}(A_k)]\log p_d^{\tau} (s_{i, k})
 + \sum_{\tau = 1}^{N_P}\big(1 - \mathbb{E}[d_{k}^{i, \tau}(A_k)]\big)\Rev{\log\left(1 - p_d^{\tau} (s_{i, k})\right)}.
\end{split}
\end{equation}
See \textbf{Appendix C} for the detailed derivations of Eq.~(\ref{26-a}).

Recall that $d_{k}^{i, \tau}(A_k)$ represents the event that target $i$ is detected via path $\tau$ in the joint association event $A_k$.
Therefore,
\begin{equation} \label{Ed}
\mathbb{E}[d_{k}^{i, \tau}(A_k)] = \sum_{j = 1}^{N_{k,M}} \mathbb{E}[a_k^{i, j, \tau}].
\end{equation}

From Eq.~(\ref{26-a}), the posterior PMF of target meta-state $q(s_{i, 1:K})$ can be updated via a forward-backward algorithm \cite{Rabiner1989}.
Note that $\xi_{i, k}$ integrates the detection probability $p_d^{\tau}, \tau = 1, ..., N_P$, resulting in improvement of the detection performance since information from all paths are integrated.  The calculation of $q(S_1^K)$ is summarized in \textbf{Module \ref{alg2}}.

\begin{algorithm}
	\caption{\textit{Detection}: calculation of $q(S_1^K)$} \label{alg2}
	\begin{algorithmic}[1]
		\REQUIRE $\mathbb{E}[A_1^K]$: multipath data association from \textbf{Module \ref{alg3}};
		\ENSURE $\big\{\hat {S}_1^K, N_{1:K}^X\big\}$: target meta-state;
		\FOR{each target $i = 1 : N_T$}
	\FOR{each time $k = 1: K$}
	\STATE Calculate $b_{k}(s_{i, k}) = \exp(\xi_{i, k}(s_{i, k}))$ with $\xi_{i, k}(s_{i, k})$ being given by Eq.~(\ref{26-b});
	\ENDFOR
		\STATE Initialize \textit{forward} variable $\overrightarrow{\alpha}_{1}(s_{i, 1}) = p(s_{i, 1}) b_{1}(s_{i, 1})$;
		\FOR{each time $k = 1 : K-1$}
		\STATE Update  $\overrightarrow{\alpha}_{k}(s_{i, k})$ by
					$\overrightarrow{\alpha}_{k+1}(s_{i, k + 1}) = \left[\sum\limits_{c = 0}^{1} \overrightarrow{\alpha}_{k}(c) T_s(s_{i, k}, c) \right] b_{k+1}(s_{i, k + 1})$;
		\ENDFOR
		\STATE Initialize  \textit{backward} variable $\overleftarrow{\beta}_{K}(s_{i, K}) = 1$;
		\FOR{each time $k = K -1 : 1$}
		\STATE Update  $\overleftarrow{\beta}_{k}(s_{i, k})$ by
			$\overleftarrow{\beta}_{k}(s_{i, k}) = \sum\limits_{c = 0}^{1} T_s(s_{i, k}, c) b_k(c) \overleftarrow{\beta}_{k+1}(c)$;
		\ENDFOR
		\STATE Calculate the posterior probability in terms of forward variable and backward variable by
			$q(s_{i, k}) = \dfrac{\overrightarrow{\alpha}_{k}(s_{i, k}) \overleftarrow{\beta}_{k}(s_{i, k})}{\sum\limits_{s_{i,k} = 0}^{1} \overrightarrow{\alpha}_{k}(s_{i,k}) \overleftarrow{\beta}_{k}(s_{i,k})}$;
		\ENDFOR
		\FOR{each time $k = 1 : K$}
		\STATE Initialize the number of confirmed tracks $N_k^X = 0$;
		\FOR{each target $i = 1 : N_T$}
		\STATE If $q(s_{i, k} = 1) \geq \delta_s$~(the track confirmation threshold), let $N_k^X = N_k^X + 1$;
		\ENDFOR
		\ENDFOR
	\end{algorithmic}
\end{algorithm}

\subsubsection{Calculation of $q(A_1^K)$}
By Eq.~(\ref{eq:QSA}) and Eq.~(\ref{18-a}), the dependence of multipath data association on target meta-state, and the dependence of measurement on target kinematic state and multipath data association shown in Fig.~\ref{Fig3}, for each time $k$, we rewrite $\log q(A_k)$ as
\begin{equation}\label{25-1}
\log q(A_k) \propto \mathbb{E}_{q(S_k)}\log p(A_k|S_k) - \sum_{j=1}^{N_{k, M}} a_k^{0, j}\log {V_{k}} + \sum_{j=1}^{N_{k, M}} \sum_{i=1}^{N_T} \sum_{\tau = 1}^{N_P} {a_k^{i, j, \tau}} \mathbb{E}_{q(x_{i, k})}  \log \mathcal{N}(y_{j,k}|x_{i,k}, R_{\tau, k}).
\end{equation}

Reshape $A_k$ as a column vector $A_k = \big[a_k^{0,1}, \ldots, a_k^{0,N_{k,M}}, a_k^{1,0,1}, \ldots, a_k^{1,0,N_P}, \ldots, a_k^{N_T,0,N_P}, a_k^{1,1,1}, \\ \ldots, a_k^{1,1,N_P}, \ldots, a_k^{1,N_{k,M},N_P}, \ldots, a_k^{N_T,N_{k,M},N_P} \big]^T$.
Take the exponential of both sides of Eq.~(\ref{25-1}) and rewrite it in the form of exponential family
with a parameter $\chi_{p, k}$ that has the same dimension with $A_k$.
\begin{equation}\label{26}
q(A_k; \chi_{p, k}) = \mathcal{Z}_k\exp\left(\chi_{p,k}^T A_k\right) \mathbb{I}\{A_k \in \mathcal{A}\},
\end{equation}
where $\chi_{p, k} \!\!=\!\! \big[\chi_k^{0,1}, \ldots, \chi_k^{0,N_{k,M}}, \chi_k^{1,0,1}, \ldots, \chi_k^{1,0,N_P}, \ldots,
\chi_k^{N_T,0,N_P},  \chi_k^{1,1,1},  \ldots, \chi_k^{1,1,N_P}, \ldots, \chi_k^{1,N_{k,M},N_P}, \\ \ldots, \chi_k^{N_T,N_{k,M},N_P} \big]^T$,
and
\begin{equation}\label{27}
\begin{split}
\chi_{p, k}^{0, j} =& -\log (V_{k}), \quad
\chi_{p, k}^{i, 0, \tau} =  \sum \nolimits_{s_{i,k}=0}^1 q(s_{i,k})\log \big(1-p_d^\tau(s_{i, k})\big), \\
\chi_{p, k}^{i,j,\tau} =& \sum_{s_{i,k} = 0}^1 q(s_{i, k})\log ({p_d^\tau(s_{i, k})}/(\lambda V_{k})) + \log(2\pi |R_{\tau,k}|^{-\frac{1}{2}}) \\
& - \frac{1}{2} \mathcal{D}(y_{j,k} - h_k^{\tau}(\hat x_{i,k}), R_{\tau, k}) - \frac{1}{2} \text{Tr}\left\{R_{\tau, k}^{-1}H_k^{\tau}P_{i,k}(H_k^{\tau})^T\right\} , \\ & i = 1, \ldots, N_T; \,\,j = 1, \ldots, N_{k,M}; \,\, \tau = 1, \ldots, N_P.
\end{split}
\end{equation}
In Eq.~(\ref{26}), the constant $\mathcal{Z}_k = \text{Poisson}(N_{k,M})$ is the Poisson distribution with mean $N_{k,M}$, and
$\mathcal{A}$ is the set of feasible joint multipath data association events.
See \textbf{Appendix D} for the detailed derivations of Eqs.~(\ref{26})-(\ref{27}).

Calculating the exact expectation $\mathbb{E}[A_k]$ from Eq.~(\ref{26}) requires the enumeration of all feasible target-to-measurement-to-path events, which is intractable since the number of feasible association events increases exponentially with the increasing of the numbers of targets, measurements and paths. 
Recently, the belief propagation schemes for data association have attracted much attention \cite{Segal2013, Williams2014,  Willams2016, meyer2017scalable, meyer2018message}. LBP runs belief propagation on a graph containing loops. Here, by constructing the factor graph corresponding to the triple target-to-measurement-to-path association, we extend the LBP of \cite{Turner2014} to approximate the marginal probabilities of multipath data association. Note that, as in \cite{sathyan2013multiple}, an alternative way to solve triple target-measurement-path association problem is
to reduce it to a regular (non-multipath) data association problem, with each (target, path) acting as a pseudo-target.
By formulating the reduced problem as a two-dimensional maximum weighted matching problem, the technique in \cite{Turner2014, Williams2014} can be applied immediately.
However, the constraints on the triple target-measurement-path association in this paper, i.e., equation group~(3)~(particularly the first equation),
making the reduced two-dimensional association problem not a two-dimensional maximum weighted matching problem, and therefore prohibit applying the technique in \cite{Turner2014, Williams2014} to the triple target-measurement-path association in this paper.

By expending the joint association event $A_k$, the posterior PDF $q(A_k)$ can be represented as a factor graph
\begin{equation}\label{eq:qakexpending}
\begin{split}
	q(A_k;\chi_{p, k}) \propto \prod_{i=1}^{N_T} \prod_{\tau = 1}^{N_P}f_{i, \tau}^R \prod_{j=1}^{N_{k, M}} \prod_{\tau = 1}^{N_P}f_{j, \tau}^C \prod_{i=1}^{N_T} \prod_{j=1}^{N_{k, M}} f_{i, j}^P
 \prod_{i=1}^{N_T} \prod_{j=0}^{N_{k, M}}\prod_{\tau = 1}^{N_P}f_{i,j,\tau}^E  \prod_{j=1}^{N_{k, M}}f_{0, j}^E
\end{split}
\end{equation}
with
\begin{equation}\nonumber
\begin{split}
 &f_{i, \tau}^R = \mathbb{I}\Big(\sum_{j = 0}^{N_{k, M}} a_k^{i, j, \tau} = 1\Big), f_{j, \tau}^C = \mathbb{I}\Big(\sum_{i = 1}^{N_T} a_k^{i, j, \tau} + a_k^{0,j} = 1 \Big), \\
& f_{i, j}^P = \mathbb{I}\Big(\sum_{\tau = 1}^{N_P} a_k^{i, j, \tau} = 1 \Big),
f_{i, j, \tau}^E = \exp\Big\{\chi_{p, k}^{i, j, \tau} a_k^{i, j, \tau}\Big\}, f_{0, j}^E = \exp\Big\{\chi_{p, k}^{0, j} a_k^{0, j}\Big\}.
\end{split}
\end{equation}
being factors, and $a_k^{i,j,\tau}$ being variables.

\begin{figure}[!htbp]
\centering
\includegraphics[width=0.55\textwidth]{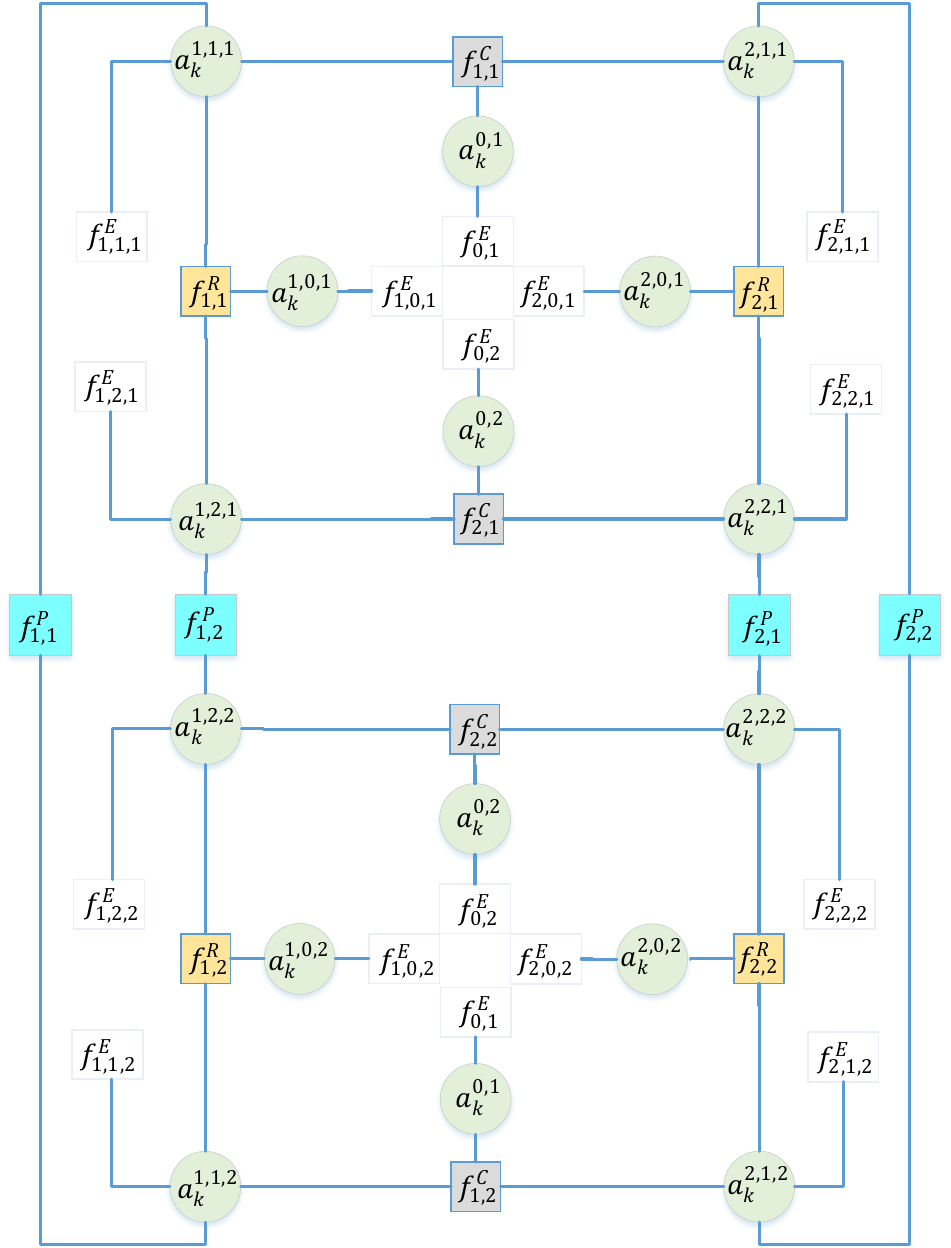}
\caption{The factor graph for modeling the joint multipath association event in Fig.~\ref{Fig2-a} (two targets, two measurements and two path): (1) a (squared) node represents a factor; (2) a (circle) node represents an individual variable of the joint association event; (3) Each factor $f$ is a function of the corresponding set of variables $a_k$. There are undirected edges connecting each factor to all of the variables on which that factor depends.}
\label{Fig5}
\end{figure}

Fig.~\ref{Fig5} provides an exemplified factor graph \cite{Loeliger2014} for modeling the association event $A_k$ in Fig.~\ref{Fig2-a}. There are four kinds of factors in factor graph, row factor $f_{i, \tau}^R$, column factor $f_{j, \tau}^C$, page factor $f_{i, j}^P$, and evidence factor $f_{i, j, \tau}^E$. We define the follow messages for LBP (omit subscript $k$ for simplicity):
\begin{equation}\nonumber
\begin{split}
\mu_{i, j, \tau}^R &:= \text{msg}_{f_{i, \tau}^R \rightarrow a^{i, j, \tau}}(x_a), \nu_{i, j, \tau}^R := \text{msg}_{a^{i, j, \tau} \rightarrow f_{i, \tau}^R}(x_a), \\
\mu_{i, j, \tau}^C &:= \text{msg}_{f_{j, \tau}^C \rightarrow a^{i, j, \tau}}(x_a), \nu_{i, j, \tau}^C := \text{msg}_{a^{i, j, \tau} \rightarrow f_{j, \tau}^C}(x_a), \\
\mu_{i, j, \tau}^P &:= \text{msg}_{f_{i, j}^P \rightarrow a^{i, j, \tau}}(x_a), \nu_{i, j, \tau}^P := \text{msg}_{a^{i, j, \tau} \rightarrow f_{i, j}^P}(x_a),
\end{split}
\end{equation}
where $\mu := \text{msg}_{f \rightarrow a}(x_a)$ denotes the message sent from factor $f$ to variable $a$  with $x_a \in \{0, 1\}$ being the state of $a$,
and $\nu := \text{msg}_{a \rightarrow f}(x_a)$ denotes the message sent from variable $a$ to factor $f$.
Using the standard update rules of BP~\cite{kschischang2001factor},
a message from a variable $a$ to a factor $f$ is
\begin{equation}\label{a-to-f}
\text{msg}_{a \rightarrow f}(x_a) = \prod_{\bar{f} \in n(a) \backslash \{f\}} \text{msg}_{\bar{f} \rightarrow a}(x_a),
\end{equation}
and a message from a factor $f$ to a variable $a$ is
\begin{equation}\label{f-to-a}
\text{msg}_{f \rightarrow a}(x_a) = \sum_{\mathbf{x} \backslash x_a } f(\mathbf{x}_f) \prod_{\bar{a} \in n(f) \backslash \{a\}} \text{msg}_{\bar{a} \rightarrow f}(x'_{\bar{a}}),
\end{equation}
where $n(a)$ is the set of neighboring factors to variable $a$, and $n(f)$ is the set of neighboring variables to factor $f$.

Message computation consists of message initialization step and message update step.
The evidence factor $f_{i, j, \tau}^E$~(leaf factor node) is utilized to initialize messages via Eq.~(\ref{M-2-1}), and the rest factors are used to update messages. Due to the fact that there is only one nonzero value in any row/column/page of $A_k$, the messages $\mu^R_{i,j,\tau}$, $\mu^C_{i,j,\tau}$, and $\mu^P_{i,j,\tau}$ are given as Eqs.~(\ref{M-2-a})-(\ref{M-2-d}), respectively.

\begin{eqnarray}\label{M-2-1}
\mu^{\text{init}}_{i,j,\tau} =
\begin{bmatrix}
f^E_{i, j, \tau}(0) \\
f^E_{i, j, \tau}(1)
\end{bmatrix}
=\begin{bmatrix} 1 \\
\exp(\chi_{p,k}^{i, j, \tau})
\end{bmatrix},  \\ \label{M-2-a}
\mu^R_{i, j, \tau}=\begin{bmatrix}
\mu^R_{i, j, \tau}(0) \\
\mu^R_{i, j, \tau}(1)
\end{bmatrix}=
 \begin{bmatrix} \sum\limits_{j_1 \neq j} \nu^R_{i, j_1, \tau}(1) \prod \limits_{j_2 \neq j, j_1} \nu^R_{i, j_2, \tau}(0) \\  \prod \limits_{j_1 \neq j} \nu^R_{i, j_1, \tau} (0) \end{bmatrix}, \label{M-2-11} \\
 \mu^C_{i, j, \tau}= \begin{bmatrix}
 \mu^C_{i, j, \tau}(0) \\
 \mu^C_{i, j, \tau}(1)
 \end{bmatrix}=
 \begin{bmatrix} \sum \limits_{i_1 \neq i} \nu^C_{i_1, j, \tau} (1) \prod \limits_{i_2 \neq i, i_1} \nu^C_{i_2, j, \tau}(0) \\  \prod \limits_{i_1 \neq i} \nu^C_{i_1, j, \tau}(0) \end{bmatrix}, \\
 \mu^P_{i, j, \tau}= \begin{bmatrix}
 \mu^P_{i, j, \tau}(0) \\
 \mu^P_{i, j, \tau}(1)
 \end{bmatrix}=\begin{bmatrix} \sum \limits_{\tau_1 \neq \tau} \nu^P_{i, j, \tau_1}(1) \prod \limits_{\tau_2 \neq \tau, \tau_1} \nu^P_{i, j, \tau_2}(0) \\  \prod \limits_{\tau_1 \neq \tau} \nu^P_{i, j, \tau_1}(0) \end{bmatrix}.  \label{M-2-d}
\end{eqnarray}

By the fact that each variable $a_k^{i,j,\tau}, i>0, j>0$ connects to factors $f^R, f^C$,  $f^P$ and $f^E$,
the messages from variable to factor $\nu^R_{i, j, \tau}, \nu^C_{i, j, \tau}$, and $\nu^P_{i, j, \tau}$  ($i > 0, j >0$) are given as
\begin{eqnarray}\label{mu-1}
\nu^R_{i, j, \tau} = \mu^C_{i, j, \tau} \cdot \mu^P_{i, j, \tau} \cdot f^E_{i, j, \tau},\\
\nu^C_{i, j, \tau} = \mu^R_{i, j, \tau} \cdot \mu^P_{i, j, \tau} \cdot f^E_{i, j, \tau},\\
\label{mu-3}\nu^P_{i, j, \tau} = \mu^R_{i, j, \tau} \cdot \mu^C_{i, j, \tau} \cdot f^E_{i, j, \tau}.
\end{eqnarray}
For $i = 0$ or $j = 0$, the factors $\nu^C_{0, j}$ and $\nu^R_{i, 0, \tau}$ only connect to the factor $f^E$, that is, $\nu^C_{0, j} = f^E_{0, j}$, $\nu^R_{i, 0, \tau} = f^E_{i, 0, \tau}$.

The messages are updated iteratively according to the factor graph. Once the iteration process terminates, the estimated marginal distribution of each variable with its expectation can be computed by multiplying the corresponding messages from adjoining factors corresponding to the variable, that is,
\begin{eqnarray}
\label{P-1}q(a_k^{i,j,\tau}) \propto
\mu^R_{i, j, \tau}\cdot\mu^C_{i, j, \tau}\cdot\mu^P_{i, j, \tau}\cdot f^E_{i, j, \tau}, \\
\label{P-2}\mathbb{E}[a_k^{i, j, \tau}] = q(a_k^{i, j, \tau} = 1).
\end{eqnarray}
See \textbf{Appendix E} for the detailed derivation of LBP for multipath data association.

The computation of $q(A_1^K)$ is summarized in \textbf{Module \ref{alg3}}.

\begin{algorithm}[h!]
\caption{\textit{Multipath data association}: Approximated by LBP}
\label{alg3}
\begin{algorithmic}[1]
\REQUIRE Measurements $Y_1^K$, target kinematic state estimation $\{\hat X_1^K, P_1^K\}$ from \textbf{Module \ref{alg1}}, target meta-state estimation $\hat S_1^K$ from \textbf{Module \ref{alg2}};
\ENSURE PMF $q(A_1^K)$, and expectation $\mathbb{E}[A_1^K]$;
\FOR{each time $k = 1 : K$}
\STATE Calculate hyper-parameter $\chi_{p,k}$ according to Eq.~(\ref{27});
\STATE Initialize messages $\mu^{\text{init}}_{i, j, \tau}$ via Eq.~(\ref{M-2-1});
\WHILE{no convergence}
\STATE Update messages $\nu^R$, $\nu^C$, $\nu^P$ via Eqs.~(\ref{mu-1}) - (\ref{mu-3});
\STATE Update messages $\!\mu^R\!, \mu^C\!, \mu^P\!$ via Eqs.~(\ref{M-2-a}) - (\ref{M-2-d});
\ENDWHILE
\STATE Calculate $q(A_k)$ and $E(A_k)$ via Eqs.~(\ref{P-1}) - (\ref{P-2});
\ENDFOR
\STATE Output $q(A_1^K)$ and $\mathbb{E}(A_1^K)$.
\end{algorithmic}
\end{algorithm}

\emph{Remark 1:}
In \cite{Williams2014}, \emph{Williams} and \emph{Lau} considered the LBP for the two-way data association problem, i.e., the correspondence between targets and measurements, in a single frame.
By modeling the two-way data association problem as a maximum weighted matching problem on a bipartite graph and showing that the message update equations are contractions, the LBP was proved to converge \cite{Williams2014}.
However, the technique in \cite{Williams2014} may not be applied to the triple target-measurement-path association that is a three-matching problem in this paper.
So the convergence of the LBP in the present paper is not guaranteed a priori.
In the Supplement Material~(Appendix A), we provide a toy example (two targets, four propagation paths, low clutter rate) to observe the convergence of the LBP algorithm.
Divergence of the LBP is not seen in extensive simulations of the toy example.
It is our future work to investigate the convergence property of the LBP for the triple target-measurement-path association.

\emph{Remark 2:}
By Eq.~(\ref{eq:qakexpending}), the factor graph that models the multipath data association $A$ with $N_T$ targets, $N_M$ measurements and $N_P$ paths has $N_TN_MN_P + N_TN_P + N_M$ variables $a^{i,j,\tau}$, $N_TN_MN_P + 2N_TN_P + N_MN_P + N_TN_M + N_M$ factors~(consisting of $N_TN_P + N_MN_P + N_TN_M$ constraint factors $f_{i,\tau}^R$, $f_{j,\tau}^C$ and $f_{i,j}^P$, and $N_TN_MN_P + N_TN_P + N_M $ evidence factor node $f_{i,j,\tau}^E$). By the fact that each variable  $a^{i,j,\tau}, i>0, j>0, \tau>0$ connects with four factors~($f_{i,\tau}^R$, $f_{j,\tau}^C$, $f_{i,j}^P$ and $f_{i,j,\tau}^E$), each $a^{i,0,\tau}, i>0, \tau > 0$ connects with two factors~($f_{i,\tau}^R$ and $f_{i,j,\tau}^E$), and each $a^{0,j}, j>0$ connects with two factors~($f_{j,\tau}^C$ and $f_{j,\tau}^E$), there are total $4N_TN_MN_P + 2N_TN_P + 2N_M$ edges in the factor graph.
For applications with a large number of targets and measurements, the gating technique can exclude the very unlikely target-to-measurement associations, reducing the number of variables and
constraint factors $f_{i,j}^R$ of the factor graph.

\Rev{\subsection{The JDT-VB Algorithm and Its Computational Complexity}}
The JDT-VB algorithm consisting of \textbf{ Modules \ref{alg1}, \ref{alg2}} and \textbf{\ref{alg3}} is summarized in TABLE.~\ref{table2}.
The iteration terminates if $\mathcal{B}(Z_1^K)$ between two consecutive  iterations are close enough or the maximum  number of iterations is reached.
\renewcommand \arraystretch{1.45}
\newcommand{\tabincell}[2]{\begin{tabular}{@{}#1@{}}#2\end{tabular}}
\begin{table} [!htbp]
	\centering
	\caption{\label{table2} The summary of JDT-VB algorithm}
	\begin{tabular}{l}
		\hline \hline
		\tabincell{l}{1: \textbf{Initialization}. Initialize $q^{0}(X_1^K)$, $q^{0}(S_1^K)$, and $N_T$.} \\ \hline
		\tabincell{l}{2: \textbf{Joint Detection and Tracking (the $r$th iteration)}\\
			\hspace{1em} (2.1) \emph{Association}: Calculate $q^{r}(A_1^{K})$ via \textbf{Module \ref{alg3}}.} \\
		\hspace{1em} (2.2) \emph{Detection}: Estimate  $q^{r}(S_1^{K})$ via \textbf{Module \ref{alg2}}. \\
		\hspace{1em} (2.3) \emph{Tracking}: Estimate  $q^{r}(X_{1}^{K})$ via \textbf{Module \ref{alg1}}. \\  \hline
		\tabincell{l}{3: \textbf{Iteration Termination.} If iteration terminates, go to \\ \hspace{1em}  step (4);  otherwise reset $r \leftarrow r + 1$  and return to (2.1).}\\ \hline
		\tabincell{l}{4: \textbf{Outputs}. Output the detection and tracking results.}\\\hline
	\end{tabular}
\end{table}

The computational complexity of JDT-VB is equal to the sum of the computational cost of Modules \ref{alg1}, \ref{alg2} and \ref{alg3},
i.e.,
\begin{equation}
c_{tot} = r_{vb} \times (c_{qx} + c_{qs} + c_{qa}),
\end{equation}
where $r_{vb}$ is the number of JDT-VB iterations, and $c_{qx}$, $c_{qs}$, $c_{qa}$ are the computational complexity of Modules \ref{alg1}, \ref{alg2} and \ref{alg3}, respectively. For Module \ref{alg1}, $c_{qx} = \mathcal{O}(2KN_TN_Pn_x^3)$~\cite{Simo2013} which is proportional to the sum of computational cost of path-dependent state estimation and multipath state fusion. Module \ref{alg2} is carried out by forward and backward algorithm,
and its computational complexity $c_{qs} = \mathcal{O}(4KN_T)$~\cite{Rabiner1989}. The LBP algorithm is used to approximate the multipath data association in Module \ref{alg3}. The main cost of LBP is the message update equation, which is $\mathcal{O}(d_a^2)$ for each variable $a$ at each iterations~\cite{chen2006data} with $d_a$ being the number of possible  values of variable $a$.
In our graphical model, there are total $N_TN_MN_P+N_TN_P + N_M$ variables, and each variable takes values of 0 and 1, i.e., $d_a = 2$. Hence, the computational cost $c_{qa} = \mathcal{O}(r_{lbp}\sum_{k=1}^{K}4N_TN_{k, M}N_P)$ with $r_{lbp}$ being the number of LBP iterations.
At each iteration, the computational cost of JDT-VB increases linearly with the number of targets, measurements and propagation paths.


\emph{Remark 3:}
There are several properties of JDT-VB:
\begin{itemize}
\item it provides an integrated solution for joint MDT for MDS in VB framework. The performance of detection and tracking is improved by the fact that multipath measurements are integrated to estimate target kinematic state and target meta-state.
\item it has a closed-loop iterative manner among multipath data association, kinematic state estimation, and meta-state estimation, which is effective in dealing with the coupling relationship between estimation errors and identification errors in the view of feedback control.
\item it has polynomial computational complexity. The multipath data association is modeled by a probabilistic graphical model, and the marginal association probabilities are calculated approximately by the LBP algorithm.
\end{itemize}

\Rev{\subsection{Initialization of JDT-VB}}
\label{SectionInit}
\begin{itemize}
\item In scan $k$, we coarsely group the measurements that have not been used to update existing tracks into different subsets $\mathcal{Y}^i_k$, $i = 1, \ldots, N_k^s$. In each subset $\mathcal{Y}^i_k$, any two measurements are within a preset threshold vector $\rho_{\tau}$ conditioned on the assumption that they are from the same target. Subsets $\mathcal{Y}^i_k$ with at least two measurements are utilized to initialize heads of new tracks by the assumption that it is rare to receive the measurement only from one path for a target under the circumstance of multipath propagation. Considering each subset $\mathcal{Y}^i_k$ with at least two measurements, for each measurement-to-path association hypothesis, we transform the measurements in $\mathcal{Y}^i_k$ from measurement space to kinematic state space, obtaining a set of the transformed kinematic states. The kinematic state estimate $\hat x_k^i$ is obtained by fusing the transformed kinematic states which have the minimum average Mahalanobis distance. The covariance $P_k^i$ is pre-determined based on measurement noise covariance. Set the initial probability $p(s_k^i = 1) = \min(1, (|\mathcal{Y}^i_k|/N_P)^2 )$.
\item In scan $k + 1$, target kinematic state estimation $\hat x_{k+1}$ and $P_{k+1}$ is performed individually by using LBP and UKF. Meanwhile, the probability $p(s_{k+1})$ is recursively updated by using the forward-backward algorithm according to the \textbf{Module \ref{alg2}}. If the average detection probability of target $i$ in three successive scans is less than the threshold $\delta_s$, the track $i$ is deleted. The measurements that do not fall into validation gates of any tracks are used to initialize new tracks. $N_T$ is the total number of confirmed temporary tracks at scan $K$.
\end{itemize}

\section{Simulation}\label{sec:simulation}
OTHR exploits the reflection effect of the ionosphere for high-frequency signal~(3-30MHz)~to detect and
track airborne/surface targets at ranges an order of magnitude greater than conventional line-of-sight radars \cite{Fabrizio2013}.
As mentioned in Section \ref{sec:introduction}, multipath propagation phenomenon often appears in OTHR \cite{pulford1998multipath}
due to the multilayer structure of the ionosphere, resulting in multiple measurements from one target.
The correspondence among targets, measurements and propagation paths is unknown.

The performance of JDT-VB is evaluated and compared against MPTF~\cite{Percival1998} and multi-detection multiple hypothesis tracker using the OTHR target tracking scenario.
For MPTF, multiple path-dependent tracks are obtained in slant coordinate by using PDA and UKF.
Then, these path-dependent tracks are fused to derive target kinematic states in the ground coordinate.
As a multi-scan joint detection and tracking algorithm for MDS, MD-MHT~\cite{sathyan2013multiple} approximates the three-dimensional multipath data association by path-dependent two-dimensional assignments and is a suboptimal but computational effective MHT algorithm.
In \cite{sathyan2013multiple}, MD-MHT is carried out based on the track-oriented framework and solved by multiple frame assignment algorithm.
We here use the hypothesis-oriented framework together with Murty's approximation method \cite{Blackman2004, cox1996efficient}, which is called MD-HMHT.
MD-PHD~\cite{Xu2015}, which is limited to track a few high-value targets due to its high computational cost~\cite{Xu2015}, is not considered here. 


\subsection{Multitarget Tracking in OTHR Scenario}
We consider an ionospheric model with two layers, E-layer and F-layer.
There are four paths in total, i.e., EE path, EF path, FE path and FF path.
Refer to TABLE.~\ref{tab:table3} for the corresponding look-up path table.
\renewcommand \arraystretch{1.3}
\begin{table} [!htbp]
	\caption{\label{tab:table3} Propagation paths}
	\centering
	\begin{tabular}{c|c|c|c|c|c}
		\hline
		\textbf{Index} & \textbf{Path} & \textbf{$\hbar_t$} & \textbf{$\hbar_r$ }& \textbf{$p_d$} & \textbf{Explanation}  \\ \hline
		$\tau =1$ & EE & $h_E$ & $h_E$ & $p_d^1$ & transmit on E and receive on E \\ \hline
		$\tau =2$ & EF & $h_E$ & $h_F$ & $p_d^2$ & transmit on E and receive on F\\ \hline
		$\tau =3$ & FE & $h_F$ & $h_E$ & $p_d^3$ & transmit on F and receive on E\\ \hline
		$\tau =4$ & FF & $h_F$ & $h_F$ & $p_d^4$ & transmit on F and receive on F\\
		\hline
	\end{tabular}
\end{table}

The OTHR measurement $y_k = [r_k, \dot r_k, \zeta_k]^T$ in slant coordinates consists of slant range $r_k$, slant range rate $\dot r_k$ and azimuth $\zeta_k$.
The target kinematic state $x_k = [g_k, {\dot g}_k, \vartheta_k, {\dot \vartheta}_k]^T$ in ground coordinates consists of ground range $g$, ground range rate $\dot g$, bearing $\vartheta$ and bearing rate $\dot \vartheta$. The measurement function is given by \cite{pulford1998multipath}
\begin{equation}\label{equ3}
\begin{split}
&r_k = r_{\alpha, k} + r_{\beta, k} \\
&\dot r_k =  \dfrac{\dot g_k}{4} \left(\dfrac{g_k}{r_{\alpha, k}} + \dfrac{g_k - d \sin(\vartheta_k)}{r_{\beta, k}}\right) \\
&\zeta_k = \arcsin\left( \dfrac {g_k \sin(\vartheta_k)}{2r_{\alpha, k}} \right)
\end{split}
\end{equation}
with
\begin{equation}\label{equ4} \nonumber
\begin{split}
r_{\alpha, k} = \sqrt{\dfrac{g_k^2}{4} + \hbar_{r}^2},  \quad
r_{\beta, k} = \sqrt{\dfrac{g_k^2-2d g_k \sin(\vartheta_k)+d^2}{4} + \hbar_{t}^2},
\end{split}
\end{equation}
where $r_{\alpha,k}$ and $r_{\beta,k}$ are the ray path of transmitter and receiver, and $d$ is the distance between the receiver and the transmitter.

\subsubsection{Scenario parameters}
The surveillance region is assumed to be $[1500, 2000]$~km in range,
$[0.428, 0.608]$~rad in azimuth, and $[-0.524, 0.524]$~km/s in range rate.
Four targets move in the surveillance region with near-constant velocity~(see Fig.~\ref{fig11-a}).
Two of them~(Target 1 and Target 2) move in parallel.
The other two~(Target 3 and Target 4) cross at $k = 20$.
Initial kinematic states and lifetime of the targets are given in TABLE.~\ref{tab:table5}.
The sampling period $T_s = 16$~s, and the number of scans $K = 30$.
The standard deviation of the slant range, Doppler and azimuth measurement errors for all paths are $\sigma_r = 5$~km,
$\sigma_{\dot r} = 0.001$~km/s, and  $\sigma_\zeta = 0.003$~rad, respectively.
The power spectral density of the process noise for targets is set to $10^{-8}$~km$^2$/s$^3$ in both $g$-direction and $\vartheta$-direction.
For ionosphere, $N_P = 4$, $\hbar_E = 100$~km and $\hbar_F = 260$~km. The distance between the transmitter and the receiver $d = 100$~km.
\renewcommand \arraystretch{1.3}
\begin{table} [!htbp]
	\caption{\label{tab:table5} Initial parameters of targets}
	\centering
	\begin{tabular}{c|c|c}
		\hline
		\textbf{Target} & \textbf{Initial state}     & \textbf{Lifetime}   \\ \hline
		$1$ & $[1700, 0.10, 0.48, 8.7\times10^{-5}]^T$         & [1, 20] \\ \hline
		$2$ & $[1750, 0.10, 0.48, 8.7\times10^{-5}]^T$         & [1, 20] \\ \hline
		$3$ & $[1850, 0.20, 0.54, 8.7\times10^{-5}]^T$         & [10, 30] \\ \hline
		$4$ & $[1915, -0.20, 0.54, 8.7\times10^{-5}]^T$     & [10, 30] \\ \hline
	\end{tabular}
\end{table}

The performance of detection and tracking algorithms is related to the number of clutter $N_c$
and the detection probability $p_d$.
Without loss of generality, we do not distinguish the detection probabilities of different paths,
i.e., $p_d^{\tau} = p_d$ for $\tau = 1,2,3,4$.
We compare the performance of JDT-VB, MPTF and MD-HMHT by varying the values of $N_c$ and $p_d$.
100 Monte Carlo runs are carried out.
All the three algorithms are implemented in MATLAB on a PC with an Intel CORE i5 CPU and 4GB RAM.

\subsubsection{Algorithm parameters}
For JDT-VB, the threshold for the convergence of $\mathcal{B}(Z_1^K)$ $\delta_T = 10^{-5}$ and the maximum number of iterations for JDT-VB $r_{max} = 20$.
Thresholds for the initialization of track heads $\rho_{\tau} = [80~\text{km}, 0.005~\text{km/s}, 0.03~\text{rad}]^T$.
The threshold for track confirmation $\delta_s = 0.85$.
The initial target meta-state probability is given by the initialization, and
transition probabilities for target meta-state are $p(0|0) = p(1|1) = 0.85$ and $p(0|1) = p(1|0) = 0.15$.
For MPTF, M/N logic rule with parameter 2/2 \& 1/3 is used for track confirmation.
For MD-HMHT, the Murty's approximation method keeps the first $n$ best hypothesis (in this paper, $n=3$ when $N_c = 125$ and $n=2$ when $N_c = 400$).
The window length of MD-HMHT is set to be three.
For both MPTF and MD-HMHT, a track will be deleted if no measurement falls into the gate of the track over three successive scans when $N_c = 125$ or five successive scans when $N_c = 400$. The gate probability $p_g = 0.971$ for all of the algorithms.

\subsection{Performance Evaluation}
\subsubsection{Performance Metrics Calculation}
The following performance metrics are used to evaluate the tracking algorithms.
\begin{itemize}
\item {Number of Valid Tracks (NVT $\uparrow$)}
\item {Track Probability of Detection (TPD $\uparrow$)}
\item {Number of False Tracks (NFT $\downarrow$)}
\item {Tentative Track Latency (TTL $\downarrow$)}
\item {Average Euclidean Error (AEE $\downarrow$)}
\item {Mean Optimal Subpattern Assignment (MOSPA $\downarrow$)} \cite{OSPA200}
\item {Total Execution Time (TET $\downarrow$)}
\end{itemize}
For the definitions of NVT, TPD, NFT, TTL, AEE, TET, the reader can refer to \cite{Gorji2011}.
Values of the metrics are computed by averaging over all Monte Carlo runs.
$\uparrow$~($\downarrow$) indicates the higher~(lower) value the metric,  the better~(worse) the performance is.
Note that tracks with minimum length five are
used to calculate the metrics.

\subsubsection{Performance comparison}
\begin{figure}[!htbp]
\centering
\subfloat[True target trajectories.]{\label{fig11-a}\includegraphics[scale=0.55]{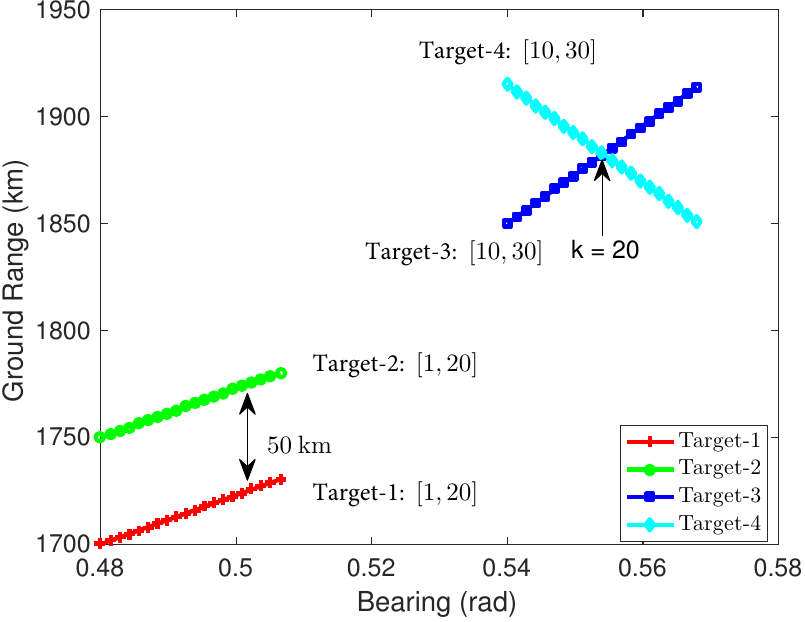}} \\
\subfloat[Multipath detection ($p_d = 0.75, N_c = 125$).]{\label{fig11-b}\includegraphics[scale=0.55]{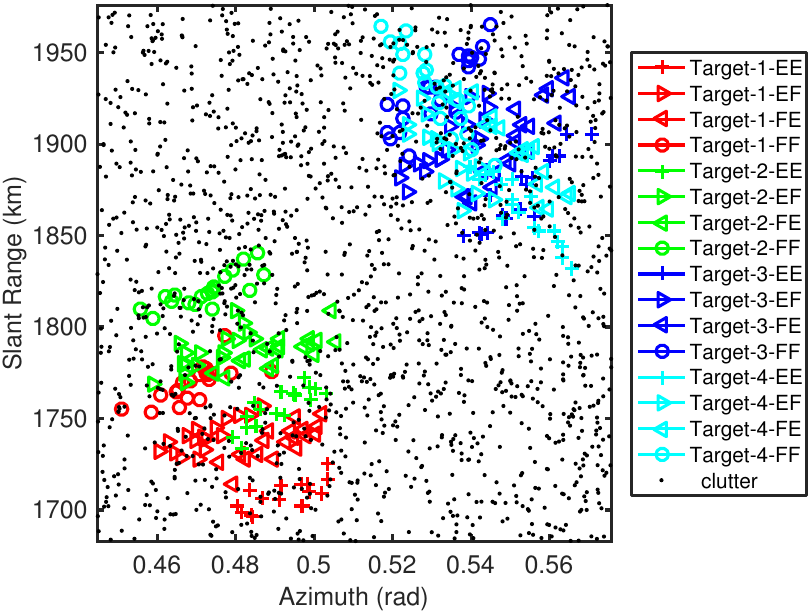}}
\caption {Target trajectories and multipath detection.}
\end{figure}
\begin{figure}[!htbp]
\centering
\subfloat[MPTF]{\label{fig12-a}\includegraphics[scale=0.55]{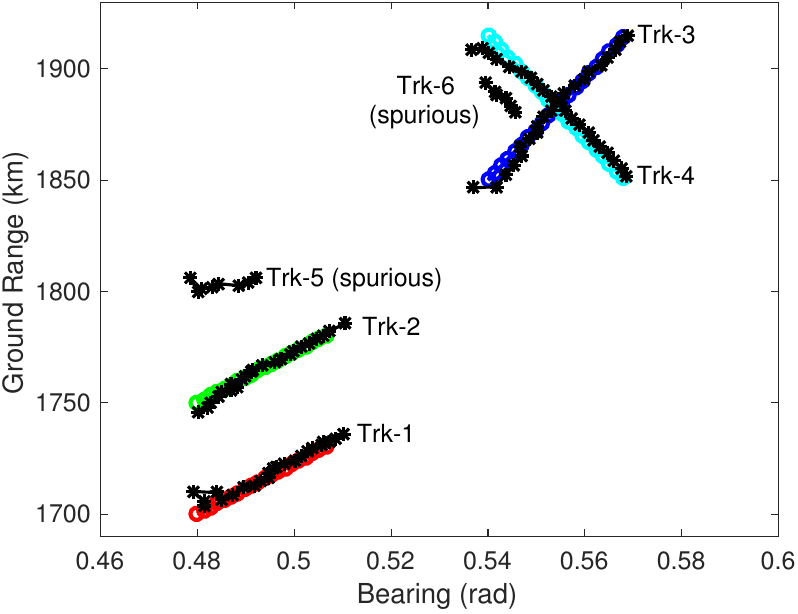}} \\
\subfloat[MD-HMHT]{\label{fig12-b}\includegraphics[scale=0.55]{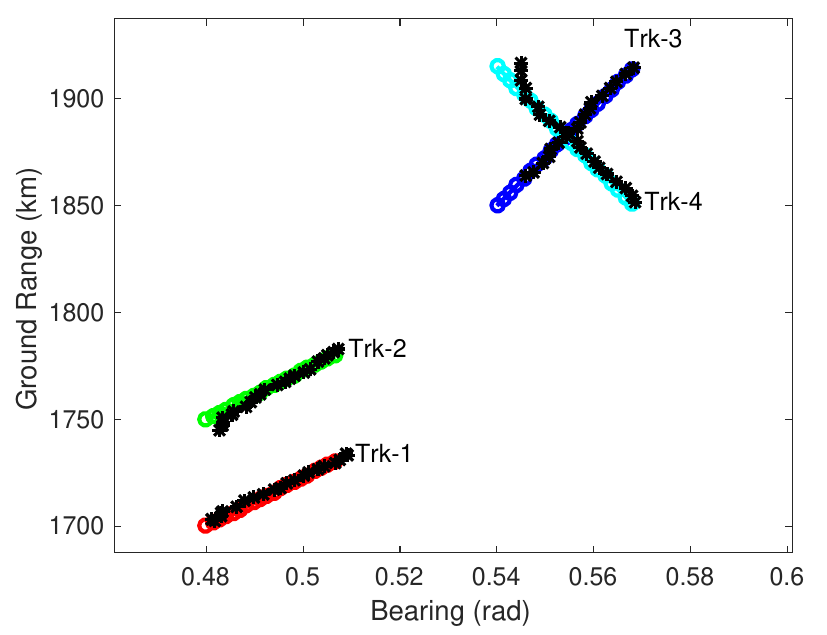}} \\
\subfloat[JDT-VB]{\label{fig12-c}\includegraphics[scale=0.55]{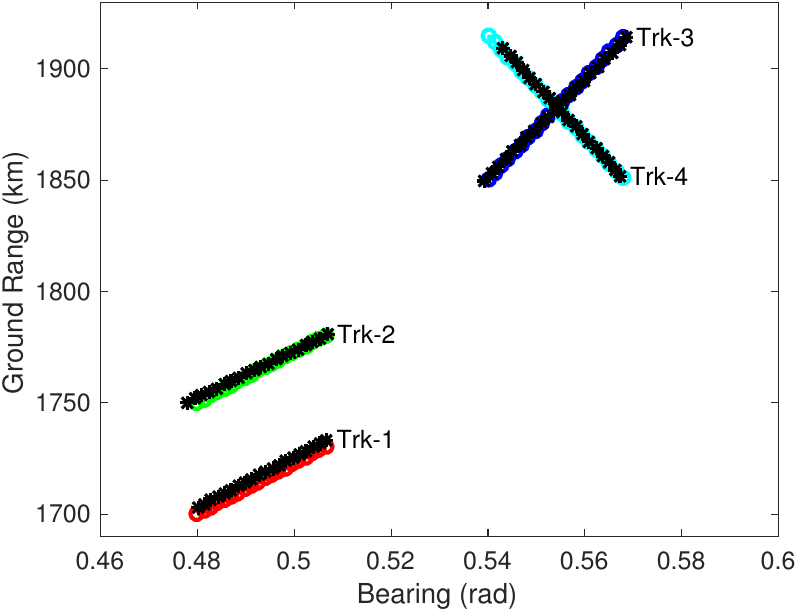}}
\caption {Tracks obtained by MPTF, MD-HMHT and JDT-VB (single run, $p_d = 0.75, N_c = 125$)}
\label{fig12}
\end{figure}

Fig.~\ref{fig11-b} shows the multipath detections of the four targets and clutter over all scans when $p_d = 0.75, N_c = 125$. The trajectories obtained by MPTF and JDT-VB in a single run~($p_d = 0.75, N_c = 125$)~are shown in Fig.~\ref{fig12}.
From Fig.~\ref{fig12-a}, it is seen that MPTF successfully tracks all four targets.
However, one false track~(Trk-5) from clutter and one ghost track (Trk-6) caused by the unsuccessful fusion of multipath tracks
corresponding to Target 4 appear as well. Fig.~\ref{fig12-b} shows that MD-HMHT successfully tracks all four targets without producing false track or ghost track.
Comparatively, as shown in Fig.~\ref{fig12-c}, the tracks for the four targets obtained by JDT-VB
are more smoother and no false track or ghost track is generated.

\begin{table*}
\renewcommand \arraystretch{1.6}
\centering
	\footnotesize
\caption{\label{tab:table10} Performance Comparison in different SNR  ($R = [5$~km, $10^{-3}$~km/s, 3$\times 10^{-3}$~rad$]$)}
\begin{center}
	\resizebox{\textwidth}{30mm}{
\begin{tabular}{c|c|c|c|c|c|c|c|c|c|c|c|c}
\hline \hline
\textbf{SNR} & \multicolumn{3}{|c}{\tabincell{c}{$p_d = 0.75$, $N_c = 125$}} & \multicolumn{3}{|c}{\tabincell{c}{$p_d = 0.75$, $N_c = 400$}} & \multicolumn{3}{|c}{\tabincell{c}{$p_d = 0.4$, $N_c = 125$}} & \multicolumn{3}{|c}{\tabincell{c}{$p_d = 0.4$, $N_c = 400$}} \\ \hline
		\textbf{Metrics} & MPTF   & MD-HMHT  & JDT-VB  & MPTF   & MD-HMHT & JDT-VB 	& MPTF    & MD-HMHT  & JDT-VB &MPTF   & MD-HMHT  & JDT-VB       \\ \hline
				NVT      & 3.92   & 3.88 	& 3.80	  & 3.84   & 3.31	& 3.72  	&2.04     & 3.04   	&3.50 	&1.88    & 1.91    & 3.23    \\ \hline
				TPD      & 0.99   & 0.96 	& 0.97	  & 0.99   & 0.86	& 0.96 		&0.63     & 0.63 	&0.88	&0.65    & 0.45    & 0.87    \\ \hline
				NFT      & 1.86   & 0.40 	& 0.57	  & 12.7   & 1.21	& 2.18 		&3.93     & 0.51  	&0.94 	&15.2    & 1.14    & 1.90    \\ \hline
				TTL      & 0.24   & 0.47 	& 0.38	  & 0.14   & 0.69   & 0.70   	&3.44     & 0.14 	&0.38	&3.58    & 0.70    & 0.48    \\ \hline
		AEE-R\tnote{1} 	 & 1.56   & 2.47	& 1.59 	  & 1.6    & 3.01	& 1.88 		&2.49     & 4.03 	&2.19	&2.47    & 4.33    & 2.41   \\ \hline
		AEE-B\tnote{1} 	 & 1.42   & 1.20    & 0.58	  & 1.54   & 1.80	& 0.50 		&2.28     & 2.10  	&0.98 	&2.24    & 2.30    & 1.16    \\ \hline
				MOSPA	 & 18.7   & 10.9	& 10.4    & 37.5   & 31.6	& 22.4  	&30.8     & 31.2 	&18.2	&39.6    & 39.7    & 23.1    \\ \hline
				TET (s)  & 9.08   & 81.8 	& 7.42	  & 58.6   & 1279	& 79.6 		&22.4     & 76.1 	&7.85	&48.5    & 1233    & 82.4   \\ \hline
	\end{tabular}}
    \begin{tablenotes}
 \footnotesize
 \item[1] AEE-R~(km) and AEE-B~(mrad) are the AEE in the direction of ground range and bearing, respectively.
 \end{tablenotes}
     \end{center}
\end{table*}

TABLE~\ref{tab:table10} provides the statistical performance of MPTF, MD-HMHT and JDT-VB
when $p_d$ takes values $0.75, 0.4$ and $N_c$ takes values $125, 400$.
As expected, the higher the SNR~(higher detection probability and less number of clutter), the better their performance is.
Specifically, in terms of NVT and TPD, JDT-VB performs better and better than MPTF and MD-HMHT with the decreasing of $p_d$ although their performances are
comparable in the case of $p_d = 0.75$. This is because JDT-VB integrates the target information from multiple propagation paths at measurement level which is more beneficial to track detection, especially in the case of low detection probability~(e.g., $p_d = 0.4$).
In the aspect of NFT, which is summation of the number of false tracks and the number of ghost tracks, the performances of JDT-VB and MD-HMHT are comparable, and are better than that of MPTF, especially when there is a large number of clutter~(e.g., $N_c = 400$). The main reason that both MD-HMHT and JDT-VB do not generate many false tracks is the adoption of the \Rev{multiple measurements clustering.}
On TTL, \Rev{JDT-VB and MD-HMHT have comparable performance,} which are worse than MPTF if $p_d = 0.75$ since it is not hard for MPTF to initialize tracks in the high detection probability case while ignoring single element subsets in the above-mentioned track initialization method for JDT-VB may generate delay for new tracks. In the low detection probability case, with the same reason as the superiority to MPTF on NVT and TPD, JDT-VB performs better.
On the tracking error, JDT-VB and MPTF have comparable performance on range; however, the former performs much better than the latter on azimuth.
MD-HMHT has the largest tracking error. The reason is that the marginal performance improvement brought by the iteration mechanism and state smooth of JDT-VB is decreasing~(Range rate is measurable but azimuth rate is not in OTHR).
The MOSPA shows that, on the whole, JDT-VB is superior to MPTF and MD-HMHT.
JDT-VB runs faster than MPTF when $N_c = 125$ but slower when $N_c = 400$. This is because the most computational expenditure of JDT-VB is caused by the multipath measurement clustering for track initialization; this expenditure grows fast with the increase of the number of measurements. When $N_c = 125$, the running time of track initialization takes 61.1\% of the total running time of JDT-VB, which increases up to 94.4\% when $N_c = 400$.
MD-HMHT is most time-consuming, especially in the case of high false-alarm~(e.g., $N_c = 400$). As a whole, JDT-VB outperforms MPTF and MD-HMHT for MDT of MDS, especially under low SNR circumstance.

\subsubsection{Performance analysis w.r.t. different iterations}

\begin{figure}[!htbp]
	\centering
	\includegraphics[scale = 0.75]{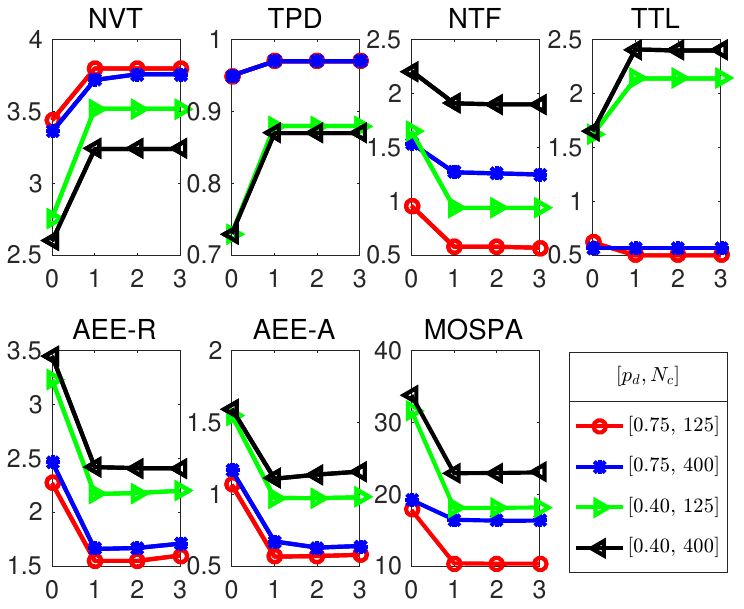}
	\caption {Performance comparison w.r.t. number of iterations.}
	\label{fig151}
\end{figure}

Fig.~\ref{fig151} shows the performance of JDT-VB w.r.t. the number of iterations considering different $p_d$ and $N_c$ for 100 Monte Carlo runs.
It is seen that JDT-VB converges quickly and its performance improves as the increasing of the number of iterations. Before the iteration starts ($r = 0$), the initial kinematic state error acquired by the multipath measurement clustering is large because of clutter and non-zero missed detection probability.
This large state error may increase the risk of incorrect target-to-measurement-to-path association
in the subsequent scans which in turn leads to large kinematic state estimation errors and/or the reduction of the track detection probability of the target.
For $r > 0$, based on the results from the last iteration, the estimations of target kinematic state and target meta-state are improved by the smooth mechanism, leading to the reduction of incorrect target-to-measurement-to-path association which in turn improves the estimation of target kinematic state and target meta-state.
By this procedure of iteration, the target detection and track performance of JDT-VB is improved significantly at the end.
More simulation results including detection and tracking for each target are given in Supplementary Material.

\section{Conclusion}\label{sec:conclusion}
We studied the MDT problem for MDS. Based on VB, we proposed a joint detection and tracking algorithm, JDT-VB, which is a closed-loop solution among multipath data association, target kinematic state estimation, and target detection. The corresponding analytical solutions are calculated iteratively via LBP, URTS, and forward-backward algorithm, respectively.
Simulations of an OTHR multitarget tracking application have shown that JDT-VB improves the performance of target detection and tracking significantly comparing with MPTF and MD-HMHT, and hence should be considered as an alternative priority for MDT of MDS.
The proposed JDT-VB algorithm was implemented off-line although it is possible to develop an online version of the algorithm by replacing the fixed-interval smoother and with filters.

\section*{Appendix}
\subsection{Derivation of $p(A_k|N_{k,M})$}
Define variables $\delta^t(A_k)$, $\delta^m(A_k)$, $\delta^{p}(A_k)$ as detection indicators of target, measurement, and path in the joint association event $A_k$, respectively. In particular, $\delta_i^t(A_k)\delta_j^m(A_k)$$\delta_{\tau}^p(A_k) = 1$ means that the $i$th target is associated with the $j$th measurement via propagation path $\tau$. Denote $N_{C}(A_k)$ the number of unassociated measurements in event $A_k$. For a given $A_k$, the vectors $\delta^t(A_k)$, $\delta^{p}(A_k)$ and $N_{C}(A_k)$ are completely defined.
Therefore,
\begin{equation} \label{100}
\begin{split}
p(A_k|N_{k,M}) =& p(A_k, \delta^t(A_k), \delta^{p}(A_k), N_C(A_k)|N_{k, M}) \\
=& p(A_k|\delta^t(A_k), \delta^{p}(A_k), N_C(A_k), N_{k, M}) p(\delta^t(A_k), \delta^{p}(A_k), N_C(A_k)|N_{k, M}).
\end{split}
\end{equation}
The first term in Eq.~(\ref{100}) is obtained from the following reasoning based on combinatorics:
\begin{itemize}
	\item In event $A_k$, there are $N_{k, M} - N_C(A_k)$ target-originated measurements and $N_C(A_k)$ false measurements.
	\item The number of joint association events $A_k$ in which the same set of the target-originated measurements is detected is given by the number of permutations of the $N_{k, M}$ measurements taken as $N_{k, M} - N_C(A_k)$.
\end{itemize}
Therefore, assuming that each such event is a priori equally likely, the first term in Eq.~(\ref{100}) is
\begin{equation} \label{100-a}
p(A_k|\delta^t(A_k), \delta^{p}(A_k), N_C(A_k), N_{k, M}) = \dfrac{N_{C}(A_k)!}{N_{k, M}!}.
\end{equation}
Assume that $\delta^t(A_k)$, $\delta^p(A_k)$ and $N_C(A_k)$ are independent.
The last term in Eq.~(\ref{100}) is,
\begin{equation} \label{100-b}
\begin{split}
p(\delta^t(A_k), \delta^{p}(A_k), N_C(A_k)|N_{k, M})=
 \mu_F(N_C(A_k))\prod_{i = 1}^{N_T} \prod_{\tau = 1}^{N_P}(p_d^\tau(i))^{\delta_i^t \delta_{\tau}^p}(1 - p_d^\tau(i))^{1- \delta_i^t \delta_{\tau}^p},
\end{split}
\end{equation}
where the PMF of the number of false measurements  $\mu_F(N_{C}(A_k))$ is
\begin{equation} \label{100-c}
\mu_F(N_{C}(A_k)) = \exp(-\lambda V_{k}) \dfrac{(\lambda V_k)^{N_{C}(A_k)}}{N_{C}(A_k)!}.
\end{equation}

Denoting $d_k^{i, \tau}(A_k) = \delta_i^t(A_k) \delta_{\tau}^p(A_k)$, and substituting Eqs.~(\ref{100-a}-\ref{100-c}) into Eq.~(\ref{100}), we obtain Eqs.~(\ref{9-E})-(\ref{Nc}).

\Rev{
\subsection{Derivation of Eq.~(\ref{B1002})}}
\Rev{
Rewrite the Gaussian PDF $\mathcal{N}(y_{j, k}|h_k^{\tau}(x_{i,k}), R_{\tau, k})$ as the following canonical form
\begin{equation}\label{eq:canonical}
\begin{split}
\mathcal{N}(y_{j, k}|h_k^{\tau}(x_{i,k}), R_{\tau, k}) =
\exp \Big[\zeta_{\tau,j,k} \!+\! \eta_{\tau, j, k}^T h_k^{\tau}(x_{i,k}) -
\dfrac{1}{2}h_k^{\tau}(x_{i,k})^T \Lambda_{\tau, k} h_k^{\tau}(x_{i,k}) \Big],
\end{split}
\end{equation}
where $\Lambda_{\tau, k} = R_{\tau, k}^{-1}$, $\eta_{\tau, j, k} = R_{\tau, k}^{-1} y_{j,k}$, and
$\zeta_{\tau, j, k} = -\dfrac{1}{2}\left(n_y\log 2\pi - \log|\Lambda_{\tau, k}| + \eta_{\tau,j,k}^T \Lambda_{\tau, k}^{-1}\eta_{\tau,j,k}\right)$.
}

\Rev{
Denote $\eta_{\tau,k} = \sum_{j=1}^{N_{k, M}} \mathbb{E}[a_k^{i, j, \tau}] \eta_{\tau,j,k}$
and $\bar{\Lambda}_{\tau,k} = \sum_{j=1}^{N_{k, M}} \mathbb{E}[a_k^{i, j, \tau}] \Lambda_{\tau, k}$,
the product of $N_{k, M}$ Gaussian PDFs $\prod_{j = 1}^{N_{k, M}} \mathcal{N}(y_{j, k}|h_k^{\tau}(x_{i,k}), R_{\tau, k})^{\mathbb{E}[a_k^{i, j, \tau}]}$  is rewritten as
\begin{equation} \label{AB1001}
\begin{split}
\prod \limits_{j = 1}^{N_{k, M}} \mathcal{N}(y_{j, k}|&h_k^{\tau}(x_{i,k}), R_{\tau, k})^{\mathbb{E}[a_k^{i, j, \tau}]} \\
&= \exp \left[\sum \limits_{j=1}^{N_{k, M}} \mathbb{E}[a_k^{i, j, \tau}] \zeta_{\tau,j,k} + \eta_{\tau,k}^T h_k^{\tau}(x_{i,k})
- \dfrac{1}{2} h_k^{\tau}(x_{i,k})^T  \bar{\Lambda}_{\tau, k} h_k^{\tau}(x_{i,k}) \right] \\
& = \exp \left[\sum \limits_{j=1}^{N_{k, M}} \mathbb{E}[a_k^{i, j, \tau}] \zeta_{\tau,j,k} - \zeta' + \zeta' + \eta_{\tau,k}^T h_k^{\tau}(x_{i,k})
-\dfrac{1}{2} h_k^{\tau}(x_{i,k})^T  \bar{\Lambda}_{\tau, k} h_k^{\tau}(x_{i,k}) \right] \\
& \propto \exp \left[\zeta' + \eta_{\tau,k}^T h_k^{\tau}(x_{i,k})
- \dfrac{1}{2} h_k^{\tau}(x_{i,k})^T  \bar{\Lambda}_{\tau, k} h_k^{\tau}(x_{i,k})  \right],
\end{split}
\end{equation}
with $\zeta' = -\dfrac{1}{2}\left(n_y\log 2\pi - \log \left|\bar{\Lambda}_{\tau, k}\right| + \eta_{\tau,k}^T \bar{\Lambda}_{\tau, k}^{-1} \eta_{\tau,k} \right)$.}

\Rev{
Based on the canonical form Eq.~(\ref{eq:canonical}) and the definitions of the synthetic measurement $\bar y_{i, \tau, k}$ and its corresponding measurement covariance $\bar R_{i, \tau, k}$ in Eq.~(\ref{eq:synold}), Eq.~(\ref{B1002}) is derived.
}

\Rev{\subsection{Derivation of $q(S_1^K)$}}
\Rev{
By the definition of Eq.~(\ref{eq:QSA}) and the reduced factorization in Eq.~(\ref{18}),
the updates of posterior PDF of each target's meta-state can be derived separately as follows.
\begin{equation}\label{eq:qs}
\log q(s_{i,1:K}) \propto \mathbb{E}_{q(x_{i, 1:K}), q(A_1^K)} \\\log p(x_{i, 1:K}, A_1^K, s_{i,1:K}, Y_1^K).
\end{equation}
Substituting Eq.~(\ref{eq:pxjoint}) into Eq.~(\ref{eq:qs}), and by
\begin{equation}
\begin{split}
\mathbb{E}_{q(x_{i, 1:K}), q(A_{1}^K)} \log p(s_{i,1:K}) & = \log p(s_{i,1:K}), \\ \nonumber
\mathbb{E}_{q(x_{i, 1:K}), q(A_{1}^K)} \log p(A_1^K | s_{i,1:K}) & = \mathbb{E}_{q(A_1^K)} \log p(A_1^K|s_{i, 1:K}), \nonumber
\end{split}
\end{equation}
yields
\begin{equation}
\log q(s_{i,1:K})  = \log p(s_{i,1:K}) + \mathbb{E}_{q(A_1^K)} \log p(A_1^K|s_{i, 1:K}) + c_s,
\end{equation}
where $c_s = \mathbb{E}_{q(x_{i, 1:K}), q(A_1^K)} \log p(Y_1^K|x_{i, 1:K}, A_1^K)p(x_{i, 1:K})$ is independent from $s_{i,1:K}$.
}

For two functions $f$ and $g$, define $f \stackrel{c}{=} g$ if $f = g + c$, where $c$ is an additive constant.
By expanding $p(A_1^K|s_{i, 1:K})$ over time $k$ and  throwing away the independent terms $c_s$, we have
\begin{equation} \label{C1001}
\log q(s_{i,1:K}) \stackrel{c}{=}  \log p(s_{i,1:K}) + \sum_{k=1}^K \mathbb{E}_{q(A_k)}\log p(A_k|s_{i, k}).
\end{equation}
Next we calculate $\mathbb{E}_{q(A_k)}\log p(A_k|s_{i, k})$.

Similar to Eq.~(\ref{9-E}), the conditional PDF $p(A_k|s_{i, k})$ is
\begin{equation} \label{C1002}
\begin{split}
p(A_k|s_{i, k}) = \dfrac{(\lambda V_k)^{N_{C}(A_k)}}{N_{k,M}!}\exp(-\lambda V_k)
 \prod_{\tau = 1}^{N_P}
(p_d^{\tau}(s_{i, k}))^{d_k^{i, \tau}(A_k)} (1 - p_d^{\tau}(s_{i, k}))^{1-d_k^{i, \tau}(A_k)},
\end{split}
\end{equation}
where the detection probability $p_d^{\tau}$ is now a function of $s_{i,k}$.

Substituting Eq.~(\ref{C1002}) into $\mathbb{E}_{q(A_k)}\log p(A_k|s_{i, k})$ and omitting the terms that are independent from $s_{i,k}$, yield,
\begin{equation} \label{C1003}
\begin{split}
\mathbb{E}_{q(A_k)}\log p(A_k|s_{i, k}) \stackrel{c}{=} \sum_{\tau = 1}^{N_P}\mathbb{E}[d_k^{i, \tau}(A_k)]\log
p_d^{\tau}(s_{i, k})
 + (1 - \mathbb{E}[d_k^{i, \tau}(A_k)]) \log (1 - p_d^{\tau}(s_{i, k})).
\end{split}
\end{equation}
Substituting Eq.~(\ref{C1003}) into Eq.~(\ref{C1001}) and taking the exponential of both sides of Eq.~(\ref{C1001}) yield Eq.~(\ref{26-a}).

\subsection{Derivation of $q(A_1^K)$}
By Eq.~(\ref{eq:QSA}) and Eq.~(\ref{18}), the updates of posterior PDF of data association can be derived separately as follows.
\begin{equation}\label{eq:qa}
\begin{split}
\log q(A_k) = \mathbb{E}_{q(S_k), q(X_k)} \Big[ \log p(X_k) + \log p(S_k)\Big] +
\mathbb{E}_{q(S_k), q(X_k)} \Big[\log p(A_k|S_k) + \log p(Y_k|X_k, A_k) \Big].
\end{split}
\end{equation}
Since
\begin{equation}
\begin{split}
\mathbb{E}_{q(S_k), q(X_k)} \log p(A_k|S_k) & = \mathbb{E}_{q(S_k)} \log p(A_k|S_k),\\ \nonumber
\mathbb{E}_{q(S_k), q(X_k)} \log p(Y_k|X_k, A_k) & = \mathbb{E}_{q(X_k)} \log p(Y_k|X_k, A_k), \nonumber
\end{split}
\end{equation}
and the first two terms of Eq.~(\ref{eq:qa}) are independent from $A_k$, we have
\begin{equation}\label{eq:qa2}
\begin{split}
\log q(A_k) \stackrel{c}{=} \mathbb{E}_{q(S_k)} \log p(A_k|S_k) + \mathbb{E}_{q(X_k)} \log p(Y_k|X_k, A_k).
\end{split}
\end{equation}
Substituting the likelihood function Eq.~(\ref{113-a}) into Eq.~(\ref{eq:qa2}), yields Eq.~(\ref{25-1}).

In the vein of Eq.~(\ref{C1002}), the PDF $p(A_k|S_k)$ is
\begin{equation} \label{D1002}
p(A_k|S_{k}) = \dfrac{(\lambda V_k)^{N_{C}(A_k)}}{N_{k,M}!}\exp(-\lambda V_k) \prod_{i = 1}^{N_T} \prod_{\tau = 1}^{N_P}
(p_d^{\tau}(s_{i, k}))^{d_k^{i, \tau}(A_k)} (1 - p_d^{\tau}(s_{i, k}))^{1-d_k^{i, \tau}(A_k)}.
\end{equation}
Substituting Eqs.~(\ref{Nc}) and (\ref{Ed}) into Eq.~(\ref{D1002}), and taking the logarithm of both sides of Eq.~(\ref{D1002}), yield
\begin{equation} \label{D1003}
\begin{split}
\log p(A_k|S_k) = N_{k,M} \log (\lambda V_{k}) - \lambda V_{k} - \log N_{k,M}! + \sum_{i = 1}^{N_T}\sum_{\tau = 1}^{N_P}a_k^{i, 0, \tau}\log (1- p_d^{\tau}(s_{i,k})) \\
+ \sum_{i = 1}^{N_T}\sum_{j=1}^{N_{k,M}}\sum_{\tau = 1}^{N_P}a_k^{i,j,\tau}\left(\log p_d^{\tau}(s_{i,k}) - \log (\lambda V_{k})\right)
\end{split}
\end{equation}

Take the exponential of both sides of Eq.~(\ref{D1003}), we have
\begin{equation} \label{D1004}
\begin{split}
p(A_k|S_k) = \mathcal{Z}_k \exp \Bigg(\sum_{i = 1}^{N_T}\sum_{\tau = 1}^{N_P}a_k^{i, 0, \tau} \log (1- p_d^{\tau}(s_{i,k}))  + \sum_{i = 1}^{N_T}\sum_{j=1}^{N_{k,M}}\sum_{\tau = 1}^{N_P}a_{k}^{i,j,\tau} \log \dfrac{p_d^{\tau}(s_{i,k})}{\lambda V_{k}}\Bigg),
\end{split}
\end{equation}
where $\mathcal{Z}_k = \text{Poisson} (N_{k,M})$ is a normalization constant.

In the vein of Eq.~(\ref{113-a}), $p(Y_k|X_k, A_k)$ is
\begin{equation}\label{D1006}
\begin{split}
p(Y_k|X_k, A_k) = \exp \Bigg( \sum \limits_{j=1}^{N_{k, M}}-a_k^{0, j} \log (V_{k}) +  \sum \limits_{i=1}^{N_T} \sum \limits_{j=1}^{N_{k, M}} \sum \limits_{\tau=1}^{N_P} a_k^{i,j,\tau} \log \mathcal{N}(y_{j,k}| h_k^{\tau}(x_{i, k}), R_{\tau, k}) \Bigg).
\end{split}
\end{equation}

Substituting Eqs.~(\ref{D1004}) and (\ref{D1006}) into Eq.~(\ref{eq:qa2}), and taking the exponential of both sides of Eq.~(\ref{eq:qa2}), yield,
\begin{equation}
\begin{split}
q(A_k; \chi_{p,k}) &= \mathcal{Z}_k \exp\Bigg(\sum_{j = 1}^{N_{k,M}}a_k^{0,j}\underbrace{(-\log(V_k))}_{\chi_{p,k}^{0,j}} + \sum_{i = 1}^{N_T}\sum_{\tau = 1}^{N_P}a_k^{i, 0, \tau} \underbrace{\mathbb{E}_{q(S_{k})}\log(1-p_d^{\tau}(s_{i,k}))}_{\chi_{p,k}^{i,0,\tau}} + \\
&\sum \limits_{i=1}^{N_T} \sum \limits_{j=1}^{N_{k, M}} \sum \limits_{\tau=1}^{N_P} a_k^{i,j,\tau}\underbrace{\Big(\mathbb{E}_{q(S_k)}\log \dfrac{p_d^{\tau}(s_{i,k})}{\lambda V_{k}} + \mathbb{E}_{q(X_k)}\log \mathcal{N}(y_{j,k}| h_k^{\tau}(x_{i, k}), R_{\tau, k})\Big)}_{\chi_{p,k}^{i,j,\tau}} \Bigg)\mathbb{I}(A_k \in \mathcal{A}) \\
                                   &= \mathcal{Z}_k \exp\left(\chi_{p, k}^T A_k\right) \mathbb{I}(A_k \in \mathcal{A}).
\end{split}
\end{equation}
The variational parameter $\chi_{p,k}$ are updated as follows.
\begin{itemize}
\item For $j = 1, \ldots, N_{k,M}$, $\chi_{p, k}^{0, j} = -\log (V_{k})$,
\item For $i = 1, \ldots, N_T; \,\, \tau = 1, \ldots, N_P$, $\chi_{p, k}^{i, 0, \tau} = \sum_{s_{i,k}=0}^1q(s_{i,k})\log (1- p_d^{\tau}(s_{i,k}))$,
\item For $i = 1, \ldots, N_T$, $j = 1, \ldots, N_{k, M}$, $\tau = 1, \ldots, N_P$
\begin{equation}
\chi_{p, k}^{i,j,\tau} =  \sum_{s_{i, k}=0}^1 q(s_{i, k})\log \left(\frac{p_d^\tau(s_{i, k})}{\lambda V_k} \right) + \underbrace{\mathbb{E}_{q(x_{i, k})}\log \mathcal{N}\left(y_{j, k}|h_k^{\tau}(x_{i, k}), R_{\tau,k}\right)}_{E_x}
\end{equation}
\end{itemize}
with
\begin{equation} \label{Ex-1}
\begin{split}
E_x =& - \frac{1}{2}\mathbb{E}_{q(x_{i, k})} \left[(y_{j,k}-h_k^{\tau}(x_{i, k}))^TR_{\tau,k}^{-1}(y_{j,k}-h_k^{\tau}(x_{i, k})) \right] + \log\left(2\pi |R_{\tau, k}|^{-\frac{1}{2}}\right) \\
=&-\frac{1}{2} \mathbb{E}_{q(x_{i, k})}\left[\text{Tr}\left\{R_{\tau,k}^{-1}\Big((y_{j,k}-h_k^{\tau}(x_{i, k}))(y_{j,k}-h_k^{\tau}(x_{i, k}))^T\Big)\right\}\right] + \log\left(2\pi |R_{\tau, k}|^{-\frac{1}{2}}\right)  \\
= &-\frac{1}{2} \mathbb{E}_{q(x_{i, k})}\left[\text{Tr}\left\{R_{\tau,k}^{-1}\left(y_{j,k}y_{j,k}^T - y_{j, k}(h_k^{\tau}(x_{i, k}))^T -y_{j, k}^Th_k^{\tau}(x_{i, k}) +  h_k^{\tau}(x_{i, k})(h_k^{\tau}(x_{i, k}))^T \right)\right\}\right] \\
&+ \log\left(2\pi |R_{\tau, k}|^{-\frac{1}{2}}\right)
\end{split}
\end{equation}
Let $x_{i, k} = \hat x_{i, k} + \bar x_{i, k}$ with $\bar x_{i, k}$ being the state estimation error.
Assume that $\mathbb{E}_{q(x_{i, k})} [\bar x_{i, k}] = 0$, i.e., $\hat x_{i,k}$ is an unbiased estimation of $x_{i, k}$.
Let $h_k^{\tau}(\hat x_{i, k} + \bar x_{i, k}) \approx h_k^{\tau}(\hat x_{i, k}) + H_k^{\tau}\bar x_{i, k}$,
where $H_k^{\tau}$ is the Jacobian matrix of the function $h_k^{\tau}$. Eq.~(\ref{Ex-1}) is then expanded as
\begin{equation}\label{Ex-2}
\begin{split}
E_x \approx &-\frac{1}{2}\underbrace{\text{Tr}\left\{R_{\tau,k}^{-1}\left( y_{j,k}y_{j,k}^T - y_{j, k}(h_k^{\tau}(\hat x_{i, k}))^T -y_{j, k}^Th_k^{\tau}(\hat x_{i, k}) + h_k^{\tau}(\hat x_{i, k}) (h_k^{\tau}(\hat x_{i, k}))^T\right)\right\}}_{\mathcal{D}( y_{j,  k}-h_k^{\tau}(\hat x_{i, k}), R_{\tau,k})} \\
&- \frac{1}{2}\text{Tr}\left\{R_{\tau,k}^{-1}(H_k^{\tau} \underbrace{\mathbb{E}_{q(x_{i, k})}[\bar x_{i, k}\bar x_{i, k}^{T}]}_{P_{i, k}} (H_k^{\tau})^T) \right\} + \log\left(2\pi |R_{\tau, k}|^{-\frac{1}{2}}\right) \\
=&-\frac{1}{2} \mathcal{D}( y_{j, k}-h_k^{\tau}(\hat x_{i, k}), R_{\tau,k}) - \frac{1}{2}\text{Tr}\left\{R_{\tau,k}^{-1}\left(H_k^{\tau}P_{i, k}(H_k^{\tau})^T\right)\right\} + \log\left(2\pi |R_{\tau, k}|^{-\frac{1}{2}}\right).
\end{split}
\end{equation}
This ends the derivations of Eqs.~(\ref{26}) and (\ref{27}).

\Rev{\subsection{LBP Derivation for Multipath Data Association}
According to the standard update rules of Eq.~(\ref{f-to-a}), we have the message $\mu_{i, j, \tau}^R$ as
\begin{equation} \label{mu-R-1}
\begin{split}
\mu^R_{i, j, \tau} = \sum_{a^{i,0,\tau}} \cdots \sum_{a^{i,j-1,\tau}} \sum_{a^{i,j+1,\tau}} \cdots  \sum_{a^{i, N_{k,M},\tau}} f_{i,\tau}^R
\prod_{\substack{j_1 = 0, \\ j_1 \neq j}}^{N_{k,M}}\nu^R_{i, j_1, \tau}.
\end{split}
\end{equation}
Recall that there is only one nonzero value in any row of $A$, i.e., if $a^{i,j,\tau} = 1$, then $a^{i, j_1, \tau} = 0$, $j_1 = 0, \ldots, j-1, j+1, \ldots, N_{k,M}$. Eq.~(\ref{mu-R-1}) can be rewritten as
\begin{equation}\label{mu-R}
	\mu^R_{i, j, \tau} \!=\! \begin{bmatrix}
	\mu^R_{i, j, \tau}(0) \\
	\mu^R_{i, j, \tau}(1)
	\end{bmatrix} \!=\! \begin{bmatrix}
	\sum \limits_{j_1 \neq j} \nu_{i, j_1, \tau}^R(1) \prod \limits_{j_2 \neq j_1, j} \nu_{i, j_2, \tau}^R(0) \\
	\prod \limits_{j_1 \neq j}\nu_{i, j_1, \tau}^R(0)
	\end{bmatrix}.
\end{equation}
}

\Rev{
Since each variable $a^{i,j,\tau}$ connects to the factors $f^E$, $f^R$, $f^C$ and $f^P$,
the message $\nu_{i, j, \tau}^R$ in Eq.~(\ref{mu-R}) is given by
\begin{equation}
\nu_{i, j, \tau}^R = \mu_{i, j, \tau}^C \mu_{i, j, \tau}^P f_{i, j, \tau}^E.
\end{equation}
Define the ratio $\tilde{\mu}_{i, j, \tau}^R$ as
\begin{equation}
\begin{split}
\tilde{\mu}_{i, j, \tau}^R :=& \dfrac{\mu_{i, j, \tau}^R(0)}{\mu_{i, j, \tau}^R(1)} = \dfrac{\sum \limits_{j_1 \neq j} \nu_{i, j_1, \tau}^R(1) \prod \limits_{j_2 \neq j_1, j} \nu_{i, j_2, \tau}^R(0)}{\prod \limits_{j_1 \neq j}\nu_{i, j_1, \tau}^R(0)} \\
=& \dfrac{\sum \limits_{j_1 \neq j} \nu_{i, j_1, \tau}^R(1) \prod \limits_{j_2 \neq j} \nu_{i, j_2, \tau}^R(0) / \nu_{i, j_1, \tau}^R(0)}{\prod \limits_{j_1 \neq j}\nu_{i, j_1, \tau}^R(0)} \\
=& \sum \limits_{j_1 = 0}^{N_{k,M}} \dfrac{ \nu_{i, j_1, \tau}^R(1)}{\nu_{i, j_1, \tau}^R(0)} - \dfrac{\nu_{i, j, \tau}^R(1)}{\nu_{i, j, \tau}^R(0)} \\
=& \sum \limits_{j_1 = 0}^{N_{k,M}} \tilde{\nu}_{i, j_1, \tau}^R - \tilde{\nu}_{i, j, \tau}^R
\end{split}
\end{equation}
with
\begin{equation} \label{t-nu-R}
\begin{split}
\tilde{\nu}_{i, j, \tau}^R := \dfrac{\tilde{\nu}_{i, j, \tau}^R(1)}{\tilde{\nu}_{i, j, \tau}^R(0)} = \dfrac{\mu_{i, j, \tau}^C(1) \mu_{i, j, \tau}^P(1) f_{i, j, \tau}^E(1)}{\mu_{i, j, \tau}^C(0) \mu_{i, j, \tau}^P(0) f_{i, j, \tau}^E(0)} = \dfrac{\exp(\lambda_{p}^{i, j, \tau})}{\tilde{\mu}_{i, j, \tau}^C\tilde{\mu}_{i, j, \tau}^P}.
\end{split}
\end{equation}
}

\Rev{
We symmetrically apply Eqs.~(\ref{mu-R})-(\ref{t-nu-R}) to obtain the messages $\mu_{i, j, \tau}^C$, $\nu_{i, j, \tau}^C$, $\tilde{\mu}_{i, j, \tau}^C$, $\tilde{\nu}_{i, j, \tau}^C$ and $\mu_{i, j, \tau}^P$, $\nu_{i, j, \tau}^P$, $\tilde{\mu}_{i, j, \tau}^P$, $\tilde{\nu}_{i, j, \tau}^P$ as follows.
\begin{eqnarray}
\mu_{i, j, \tau}^C = \begin{bmatrix}
\sum \limits_{i_1 \neq i} \nu_{i_1, j, \tau}^C(1) \prod \limits_{i_2 \neq i_1, i} \nu_{i_2, j, \tau}^C(0) \\
\prod \limits_{i_1 \neq i}\nu_{i_1, j, \tau}^C(0)
\end{bmatrix}, \quad
\nu_{i, j, \tau}^C = \mu_{i, j, \tau}^R \mu_{i, j, \tau}^P f_{i, j, \tau}^E, \\
\tilde{\mu}_{i, j, \tau}^C = \sum \limits_{i_1 = 0}^{N_T} \tilde{\nu}_{i_1, j, \tau}^C - \tilde{\nu}_{i, j, \tau}^C, \quad
\tilde{\nu}_{i, j, \tau}^C = \dfrac{\exp(\lambda_{p}^{i, j, \tau})}{\tilde{\mu}_{i, j, \tau}^R\tilde{\mu}_{i, j, \tau}^P},
\end{eqnarray}
and
\begin{eqnarray}
\mu_{i, j, \tau}^P = \begin{bmatrix}
\sum \limits_{\tau_1 \neq \tau} \nu_{i, j, \tau_1}^P(1) \prod \limits_{\tau_2 \neq \tau_1, \tau} \nu_{i, j, \tau_2}^P(0) \\
\prod \limits_{\tau_1 \neq \tau}\nu_{i, j, \tau_1}^P(0)
\end{bmatrix}, \quad
\nu_{i, j, \tau}^P = \mu_{i, j, \tau}^R \mu_{i, j, \tau}^C f_{i, j, \tau}^E, \\
\tilde{\mu}_{i, j, \tau}^P = \sum \limits_{\tau_1 = 1}^{N_P} \tilde{\nu}_{i_1, j, \tau}^P - \tilde{\nu}_{i, j, \tau}^P, \quad
\tilde{\nu}_{i, j, \tau}^P = \dfrac{\exp(\lambda_{p}^{i, j, \tau})}{\tilde{\mu}_{i, j, \tau}^R\tilde{\mu}_{i, j, \tau}^C}.
\end{eqnarray}
}

\Rev{
Finally, the marginal distribution of multipath data association $\mathbb{E}[a^{i, j, \tau}]$ is computed by normalizing the product of the incoming messages to each variable, i.e.,
\begin{equation}
\begin{split}
\mathbb{E}[a^{i, j, \tau}] =& \dfrac{p(a^{i, j, \tau} = 1)}{p(a^{i, j, \tau} = 1) + p(a^{i, j, \tau} = 0)}
= \dfrac{1}{1 + \dfrac{p(a^{i, j, \tau} = 1)}{p(a^{i, j, \tau} = 0)}} \\
=&\dfrac{1}{1 + \dfrac{\mu_{i, j, \tau}^R(1)\mu_{i, j, \tau}^C(1)\mu_{i, j, \tau}^P(1)f_{i,j, \tau}^E(1)}{\mu_{i, j, \tau}^R(0)\mu_{i, j, \tau}^C(0)\mu_{i, j, \tau}^P(0)f_{i,j, \tau}^E(0)}} \\
=& \dfrac{1}{1 + \exp(-x)}
\end{split}
\end{equation}
with
\begin{equation}
\begin{split}
x := \log \left(\dfrac{\mu_{i, j, \tau}^R(0)\mu_{i, j, \tau}^C(0)\mu_{i, j, \tau}^P(0)f_{i,j, \tau}^E(0)}{\mu_{i, j, \tau}^R(1)\mu_{i, j, \tau}^C(1)\mu_{i, j, \tau}^P(1)f_{i,j, \tau}^E(1)}\right)  \\
= \log \tilde{\mu}_{i, j, \tau}^R + \log \tilde{\mu}_{i, j, \tau}^C + \log \tilde{\mu}_{i, j, \tau}^P - \lambda_{p}^{i, j, \tau}.
\end{split}
\end{equation}}
\bibliographystyle{IEEEtran} 
\bibliography{IEEEabrv,OTHR-VB} 

\end{document}


\begin{center}
  \bf {\Large {Supplementary Material for: \\ Joint Target Detection and Tracking in Multipath Environment: A Variational Bayesian Approach}}
  	
  	\bigskip
  	Hua Lan, Shuai Sun, Zengfu Wang, Quan Pan, Zhishan Zhang
  	
\end{center}

\noindent

\hrule
\bigskip
\bigskip

This supplementary document covers a toy example of the multipath data association to observe the convergence of the LBP and
additional simulation results.

\section{A toy example of the multipath data association}
Here we give a toy example of the multipath data association to numerically observe the convergence of the LBP.
In this toy example, we use the OTHR target tracking scenario again.
The settings of the scenario and the relevant parameters are the same as in the paper if they are not explicitly stated.
Four different scenario cases each of which has different number of targets $N_T$, distance $\Delta R$ between the two targets in ground range,
detection probability $p_d$ and number of clutter $N_c$ that falls into the gate of the target, are designed to observe the convergence of the LBP through 1000 Monte Carlo runs.
\begin{itemize}
\item Case 1~(iterations w.r.t $N_T$): $N_t = \{2, 4, 6, 8\}$, $\Delta R = 15~\text{km}$, $p_d = 0.8$, and $N_c = 2$.
\item Case 2~(iterations w.r.t $\Delta R$): $N_t = 2$, $\Delta R = \{5~\text{km}, 10~\text{km}, 15~\text{km}, 20~\text{km}\}$, $p_d = 0.8$, and $N_c = 2$.
\item Case 3~(iterations w.r.t $p_d$): $N_t = 2$, $\Delta R = 15~\text{km}$, $p_d = \{0.4, 0.6, 0.8, 1.0\}$, and $N_c = 2$.
\item Case 4~(iterations w.r.t $N_c$): $N_t = 2$, $\Delta R = 15~\text{km}$, $p_d = 0.8$, and $N_c = \{0, 2, 4, 6\}$.
\end{itemize}

\begin{figure}[!htbp]
	\centering
	\subfloat[Iterations w.r.t $N_T$]{\label{fig12-a}\includegraphics[scale=0.48]{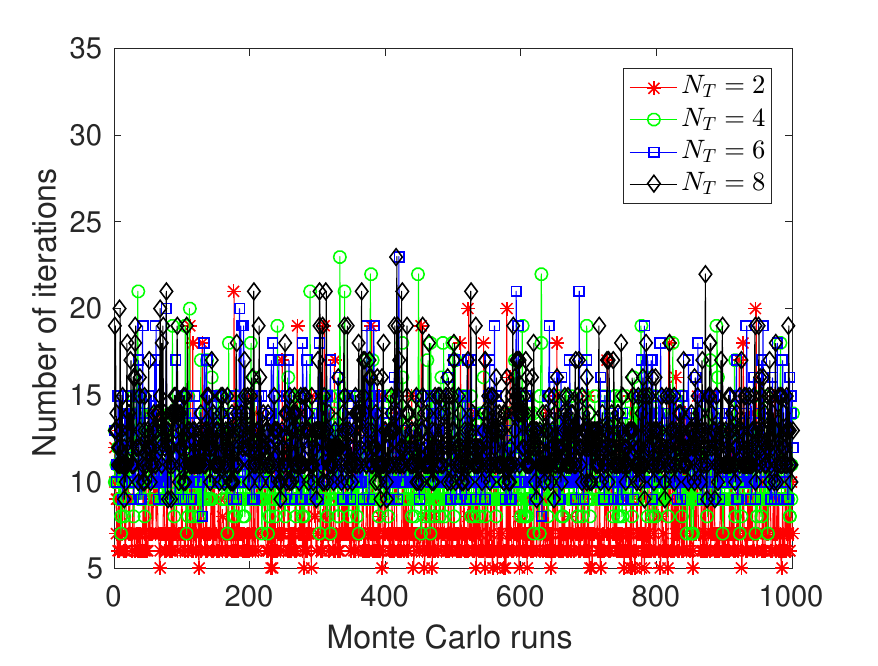}} \hspace{10pt}
	\subfloat[Iterations w.r.t $\Delta R$]{\label{fig12-b}\includegraphics[scale=0.48]{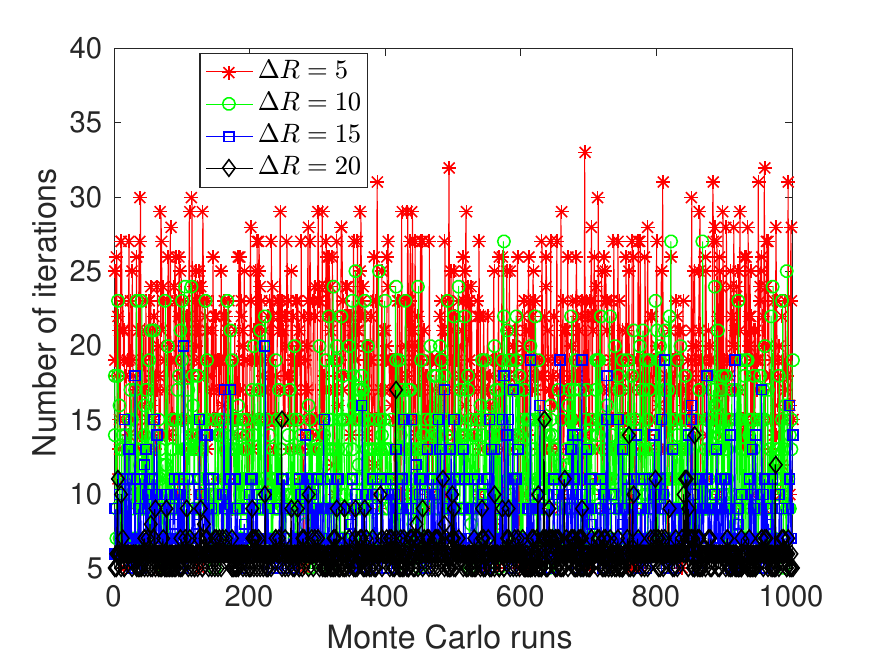}} \\
	\subfloat[Iterations w.r.t $p_d$]{\label{fig12-c}\includegraphics[scale=0.48]{Figures/Pd.pdf}}\hspace{10pt}
	\subfloat[Iterations w.r.t $p_d$]{\label{fig12-c}\includegraphics[scale=0.48]{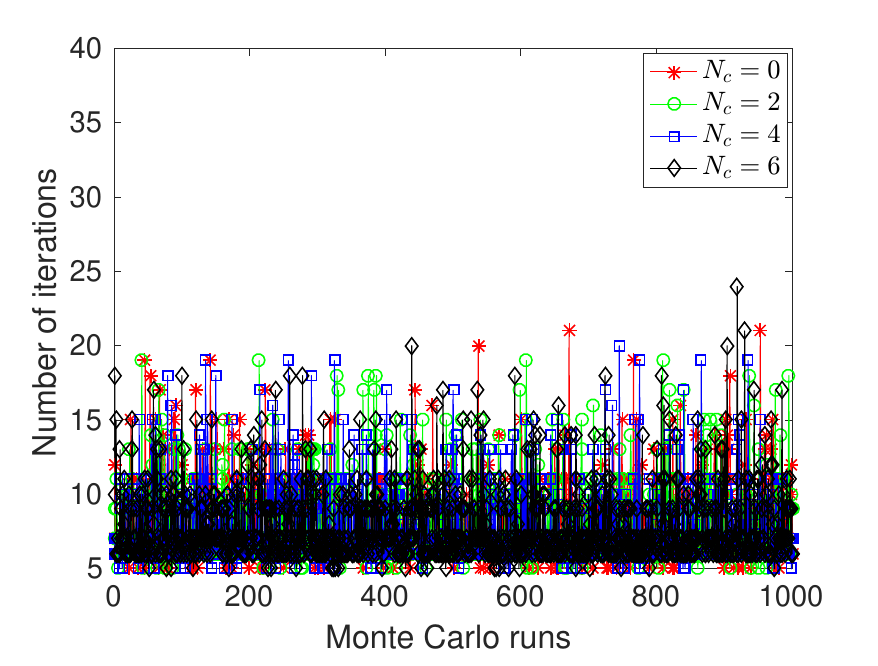}}
	\caption {Number of iterations of LBP w.r.t different cases}
\label{fig12}
\end{figure}

We set that LBP terminates if the maximum difference of the messages at two consecutive iterations is less than $10^{-5}$.
Fig.~\ref{fig12} shows the number of iterations of the LBP when it converges for the four cases.
Divergence of the LBP is not seen in these simulations of the toy example.

\section{Detection and tracking performance for each target}
\begin{figure}[!htbp]
	\centering
	\subfloat[Target 1]{\includegraphics[scale=0.54]{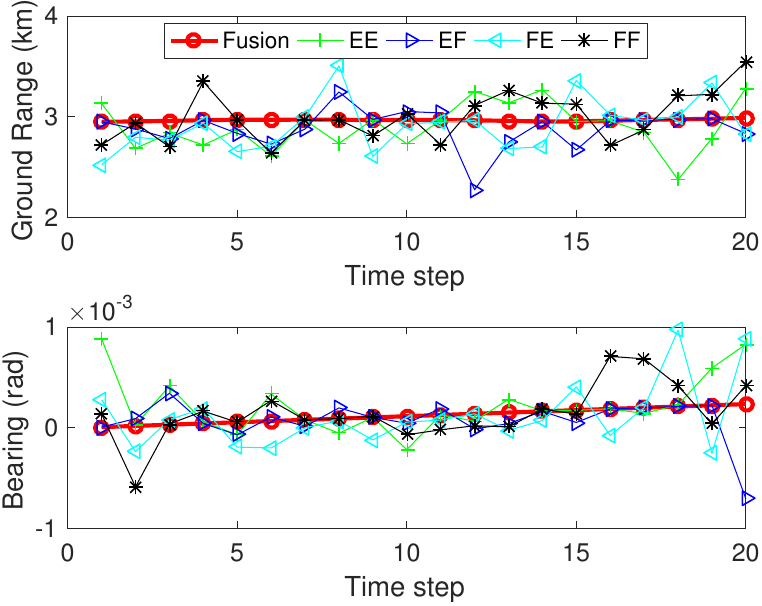}} \hspace{30pt}
	\subfloat[Target 2]{\includegraphics[scale=0.54]{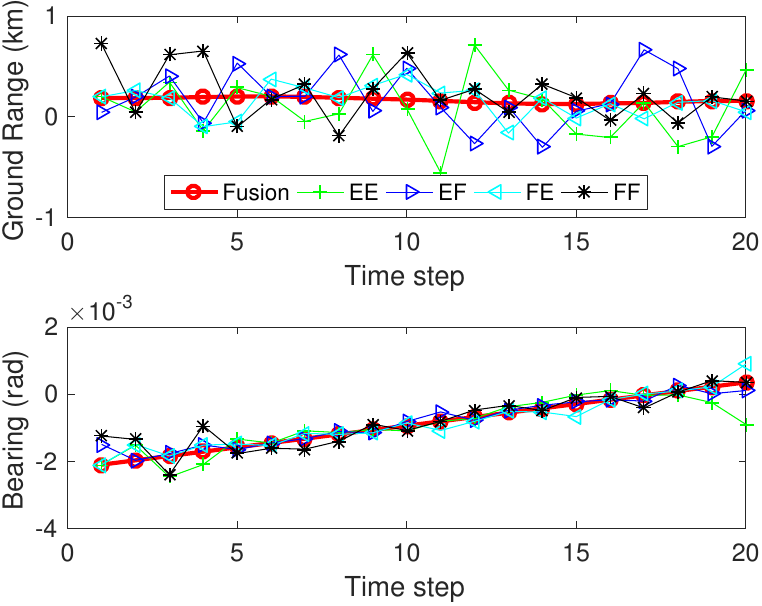}}  \\
	\subfloat[Target 3]{\includegraphics[scale=0.54]{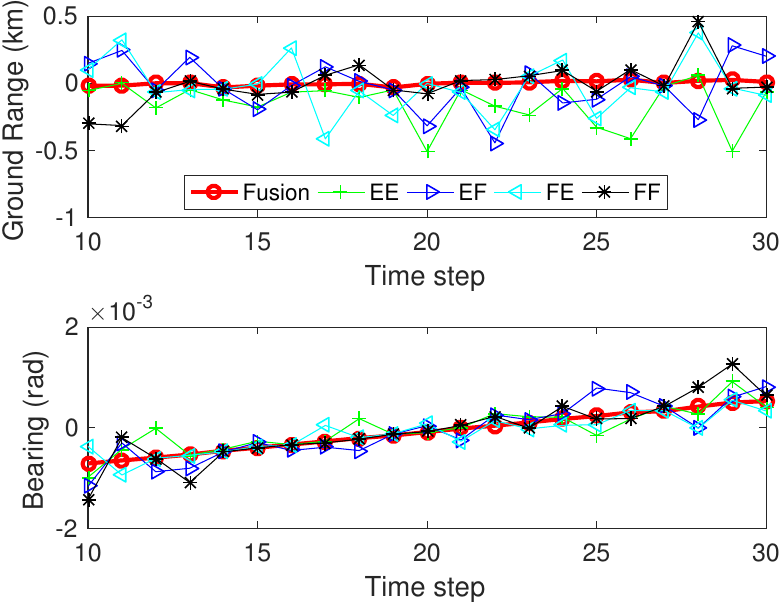}}  \hspace{20pt}
	\subfloat[Target 4]{\includegraphics[scale=0.54]{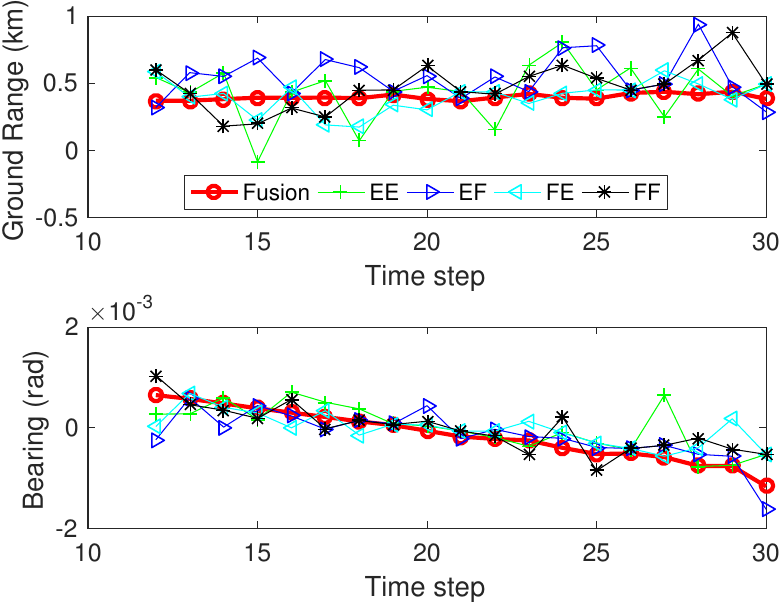}}
	\caption {Target kinematic state estimation error using each single path measurement and multipath measurements.}
    \label{fig17}
\end{figure}

\begin{figure}[!htbp]
	\centering
	\subfloat[Target 1]{\includegraphics[scale=0.54]{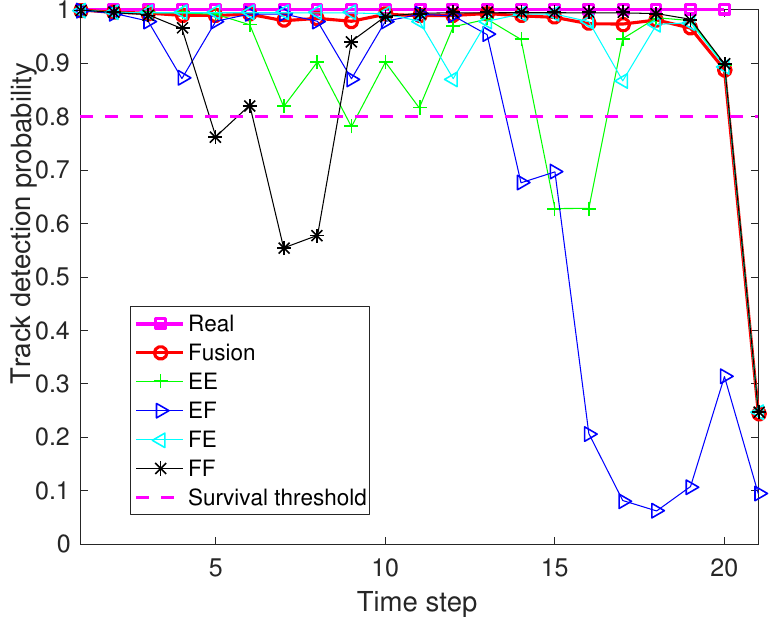}} \hspace{20pt}
	\subfloat[Target 2]{\includegraphics[scale=0.54]{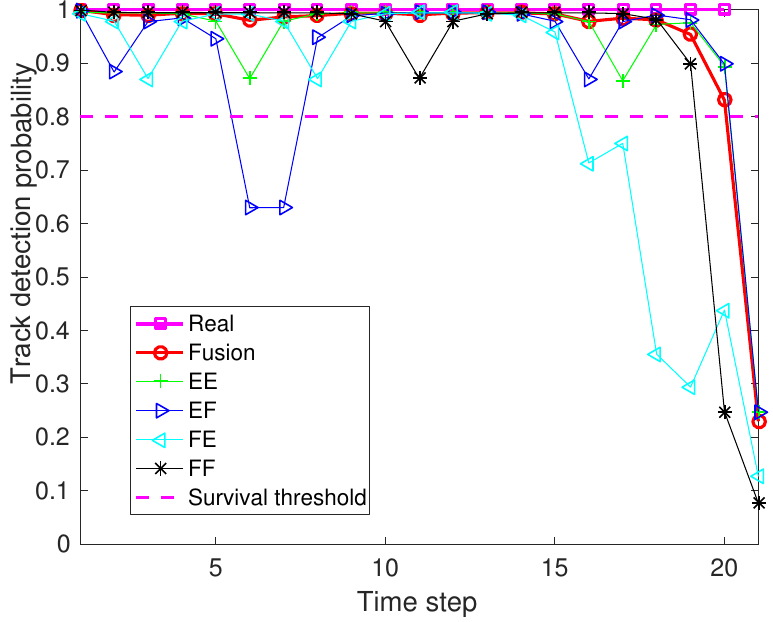}}  \\
	\subfloat[Target 3]{\includegraphics[scale=0.54]{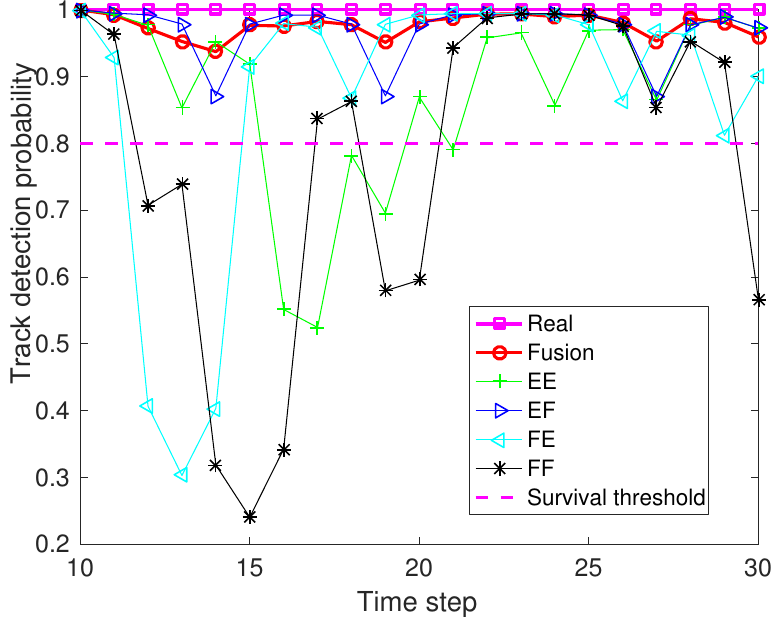}}  \hspace{20pt}
	\subfloat[Target 4]{\includegraphics[scale=0.54]{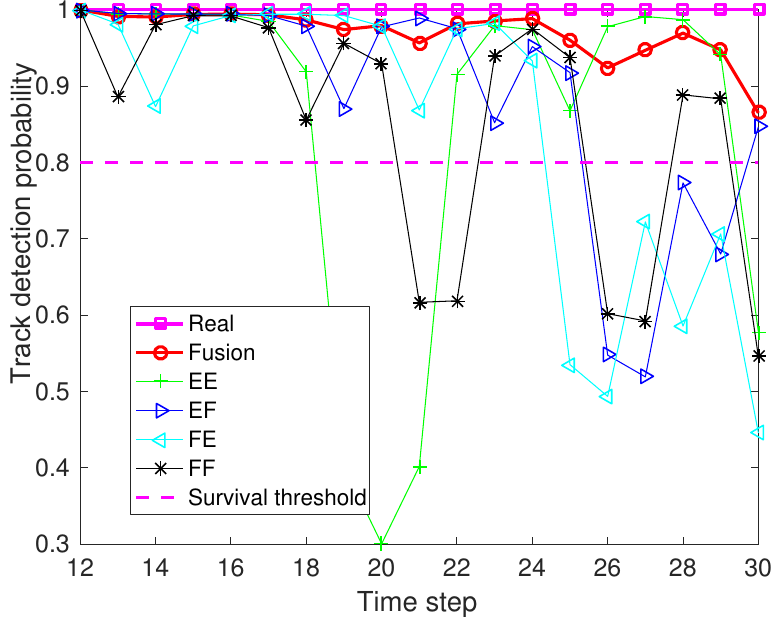}}
	\caption {Target track detection probability using each single path measurement and multipath measurements.}
    \label{fig18}
\end{figure}

\begin{figure}[!htbp]
	\centering
	\subfloat[Target 1]{\includegraphics[scale=0.54]{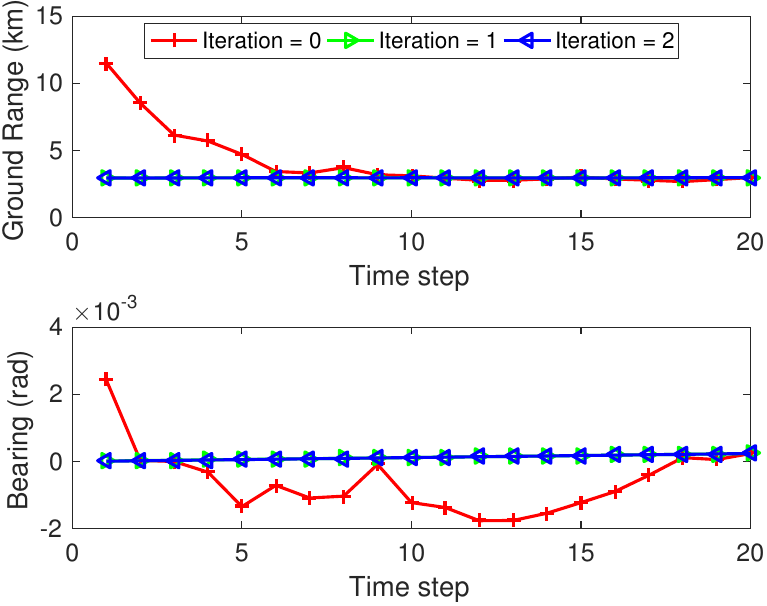}} \hspace{20pt}
	\subfloat[Target 2]{\includegraphics[scale=0.54]{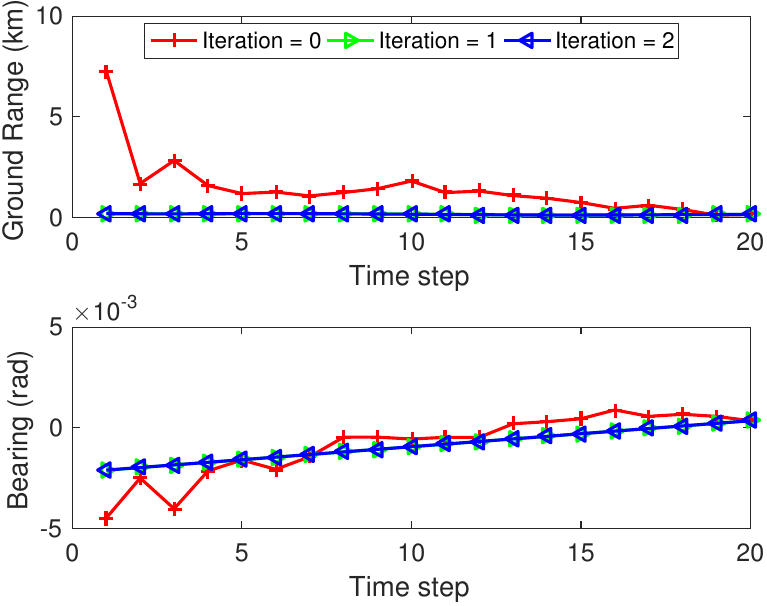}}  \\
	\subfloat[Target 3]{\includegraphics[scale=0.54]{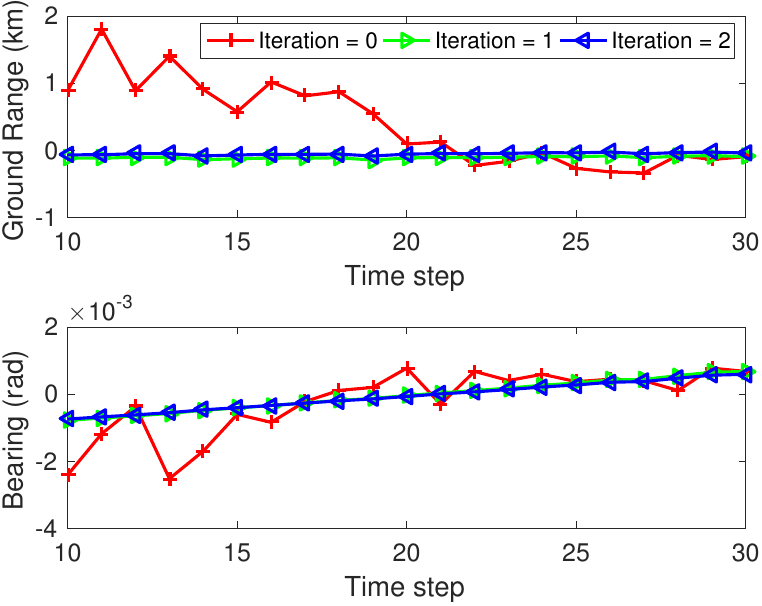}}  \hspace{20pt}
	\subfloat[Target 4]{\includegraphics[scale=0.54]{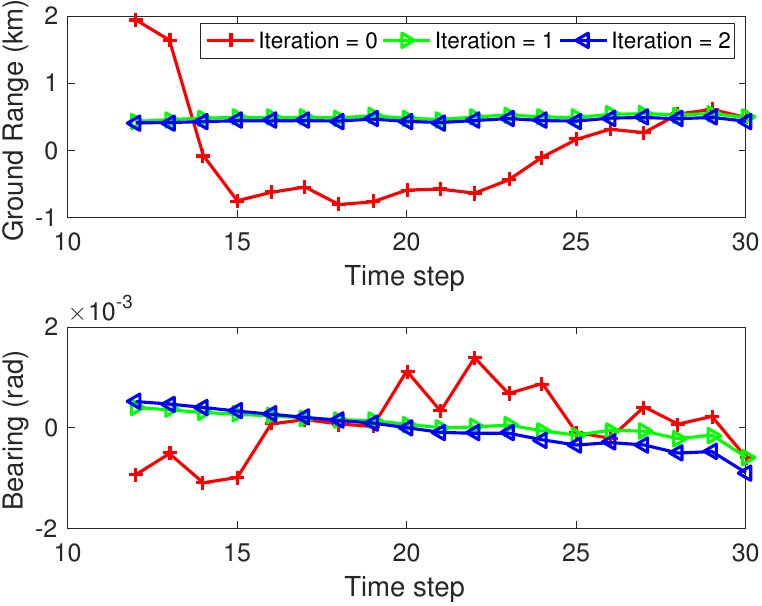}}
	\caption {Estimation error w.r.t. different number of iterations}
    \label{fig19}
\end{figure}

\begin{figure}[!htbp]
	\centering
	\subfloat[Target 1]{\includegraphics[scale=0.54]{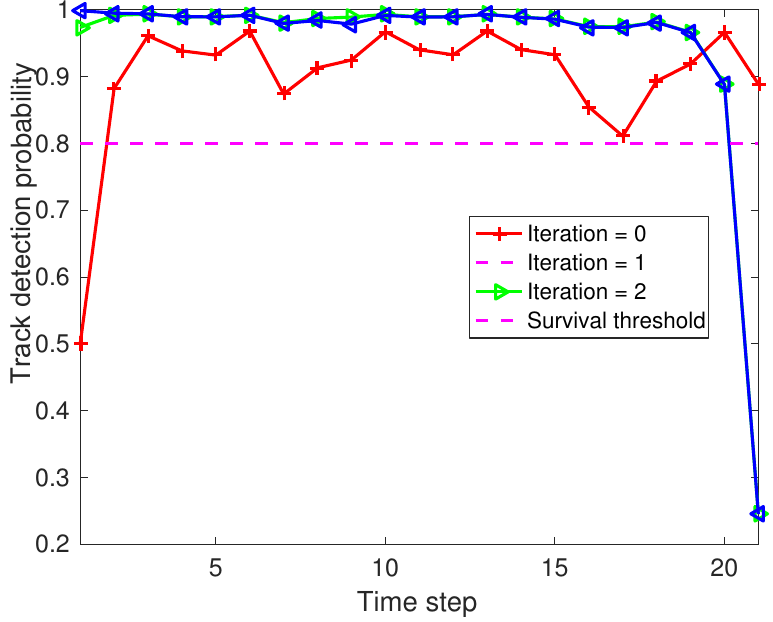}} \hspace{20pt}
	\subfloat[Target 2]{\includegraphics[scale=0.54]{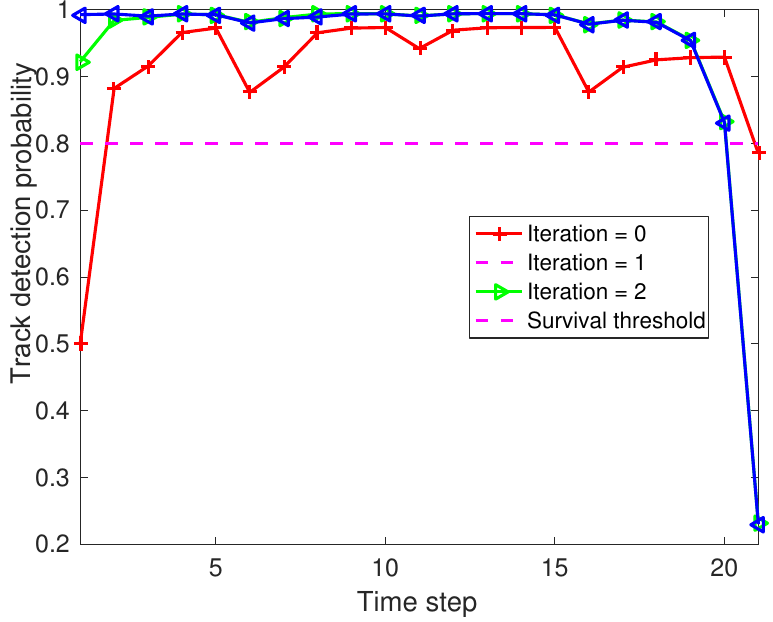}}  \\
	\subfloat[Target 3]{\includegraphics[scale=0.54]{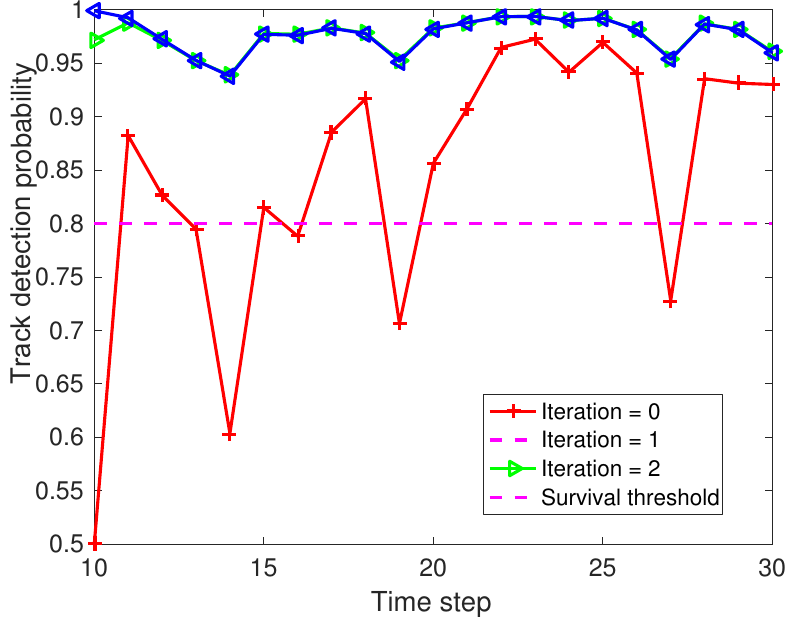}}  \hspace{20pt}
	\subfloat[Target 4]{\includegraphics[scale=0.54]{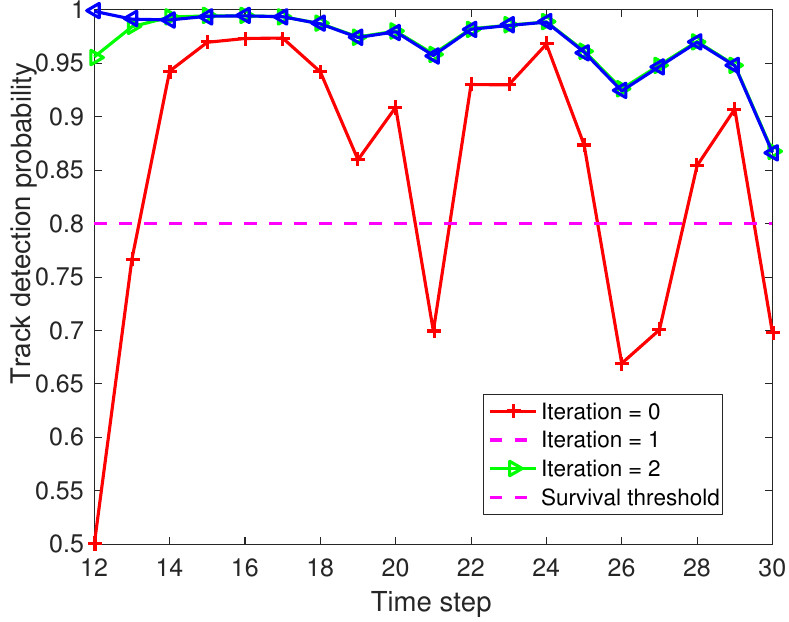}}
	\caption {Target track detection probability w.r.t. different number of iterations}
    \label{fig20}
\end{figure}

Recall that there are four targets in the scenario.
Here we present the simulation results of our proposed JDT-VB algorithm for a single run.
Fig.~{\ref{fig17}} shows the errors on ground range and bearing obtained by JDT-VB using each single path measurement and multipath measurements, while Fig.~{\ref{fig18}} shows the corresponding target track detection probability.
It is seen that the state~(including both the kinematic state and the meta-state) estimation using multipath measurements is more reliable than that using each single path measurement.
Fusing multipath measurements leads to the improvement of tracking and detection performance.
The effect of the number of iterations on the estimation and detection performance are demonstrated in Fig.~{\ref{fig19}} and Fig.~{\ref{fig20}}, respectively.
During each target's lifetime, the target kinematic state estimation error decreases and the target track detection probability increases with the increasing of the number of iterations, which illustrates the convergence of JDT-VB algorithm.
The iterative mechanism of JDT-VB is capable of reducing the performance
deterioration caused by the coupling between estimation errors
and identification errors, resulting in the improvement of tracking and detection performance.

